\documentclass{article}

\usepackage[dvipsnames]{xcolor}
\usepackage{microtype}
\usepackage{subfigure}
\usepackage{booktabs} 

\usepackage{amsmath,amsfonts,bm}

\def\eqref#1{equation~\ref{#1}}

\def\1{\bm{1}}

\DeclareMathAlphabet{\mathsfit}{\encodingdefault}{\sfdefault}{m}{sl}
\SetMathAlphabet{\mathsfit}{bold}{\encodingdefault}{\sfdefault}{bx}{n}

\usepackage[hyperfootnotes=false]{hyperref}
\usepackage{booktabs}
\usepackage{graphicx}
\usepackage{amsmath}
\usepackage{wrapfig}
\usepackage{longtable}
\usepackage{caption}
\usepackage{hyperref}

\usepackage[accepted]{arxiv}

\usepackage{amsmath}
\usepackage{amssymb}
\usepackage{mathtools}
\usepackage{amsthm}

\usepackage[capitalize,noabbrev]{cleveref}

\theoremstyle{plain}

\theoremstyle{definition}

\theoremstyle{remark}

\usepackage[textsize=tiny]{todonotes}

\icmltitlerunning{Correlating and Predicting Human Evaluations of Language Models from Natural Language Processing Benchmarks}

\begin{document}

\twocolumn[
\icmltitle{Correlating and Predicting Human Evaluations of Language Models from Natural Language Processing Benchmarks}





\begin{icmlauthorlist}
\icmlauthor{Rylan Schaeffer}{stanfordcs,note}
\icmlauthor{Punit Singh Koura}{meta}
\icmlauthor{Binh Tang}{meta}
\icmlauthor{Ranjan Subramanian}{meta}
\icmlauthor{Aaditya K. Singh}{ucl,note}
\icmlauthor{Todor Mihaylov}{meta}
\icmlauthor{Prajjwal Bhargava}{meta}
\icmlauthor{Lovish Madaan}{meta}
\icmlauthor{Niladri S. Chatterji}{meta}
\icmlauthor{Vedanuj Goswami}{meta}
\icmlauthor{Sergey Edunov}{meta}
\icmlauthor{Dieuwke Hupkes}{meta}
\icmlauthor{Sanmi Koyejo}{stanfordcs}
\icmlauthor{Sharan Narang}{meta}
\end{icmlauthorlist}

\icmlaffiliation{stanfordcs}{Stanford Computer Science}
\icmlaffiliation{ucl}{University College London}
\icmlaffiliation{meta}{Meta GenAI}
\icmlaffiliation{note}{Work completed during Meta GenAI internship.}

\icmlcorrespondingauthor{Rylan Schaeffer}{rschaef@cs.stanford.edu}
\icmlcorrespondingauthor{Sharan Narang}{sharann@meta.com}

\icmlkeywords{Machine Learning, ICML}

\vskip 0.3in
]



\printAffiliationsAndNotice{}  

\begin{abstract}
    The explosion of high-performing conversational language models (LMs) has spurred a shift from classic natural language processing (NLP) benchmarks to expensive, time-consuming and noisy human evaluations — yet the relationship between these two evaluation strategies remains hazy. In this paper, we conduct a large-scale study of four Chat Llama 2 models, comparing their performance on 160 standard NLP benchmarks (e.g., MMLU, ARC, BIG-Bench Hard) against extensive human preferences on more than 11k single-turn and 2k multi-turn dialogues from over 2k human annotators. Our findings are striking: most NLP benchmarks strongly correlate with human evaluations, suggesting that cheaper, automated metrics can serve as surprisingly reliable predictors of human preferences. Three human evaluations, such as adversarial dishonesty and safety, are anticorrelated with NLP benchmarks, while two are uncorrelated.
    Moreover, through overparameterized linear regressions, we show that NLP scores can accurately predict human evaluations across different model scales, offering a path to reduce costly human annotation without sacrificing rigor. Overall, our results affirm the continued value of classic benchmarks and illuminate how to harness them to anticipate real-world user satisfaction — pointing to how NLP benchmarks can be leveraged to meet evaluation needs of our new era of conversational AI.
\end{abstract}

\section{Introduction}

For decades, the field of natural language processing (NLP) has relied on academic benchmarks and automated metrics (e.g., Accuracy, Brier Score \citep{brier1950verification}, BLEU \cite{papineni2002bleu}) to evaluate the performance of language models (LMs). These NLP benchmarks provide a standardized and efficient way to measure model capabilities such as machine translation, text summarization, and question answering \citep{wang2018glue,wang2019superglue,srivastava2022beyond,eleuther2023evalharness,wang2023decodingtrust}. However, the recent emergence of highly capable chat LMs such as GPT \citep{ouyang2022training,openai2023gpt4}, Llama \citep{touvron2023llama1, touvron2023llama2,dubey2024llama3herdmodels}, Gemini \citep{team2023gemini,reid2024gemini} and Claude \citep{anthropic2023claude3} has prompted a re-evaluation of how we assess LMs, with a growing emphasis on assessing LMs based on their ability to interact with and assist human users in real-world scenarios \citep{zheng2023lmsyschat1m,reuel2024openproblemstechnicalai}.

\begin{figure*}[t!]
    \centering
    \includegraphics[width=0.9\linewidth]{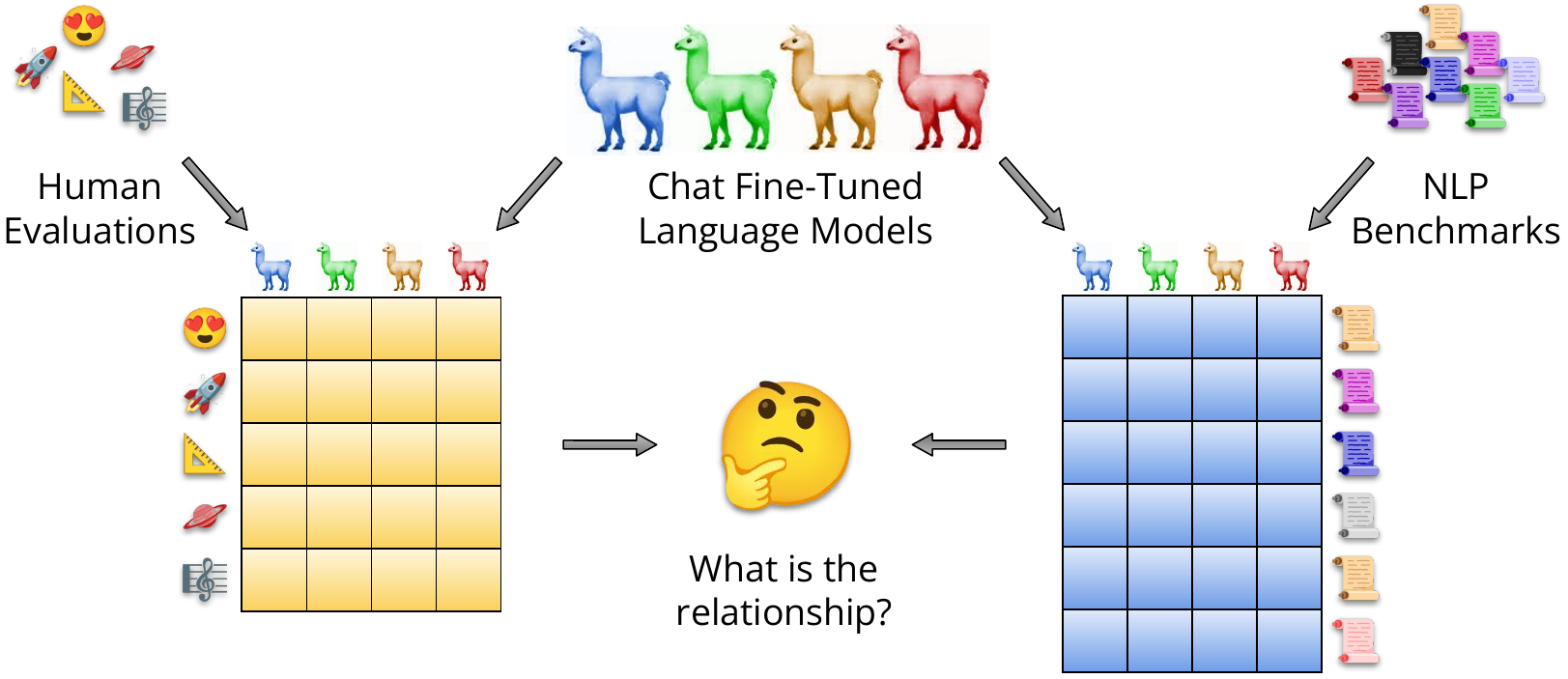}
    \caption{\textbf{Correlating and Predicting Human Evaluations of Language Models from Natural Language Processing (NLP) Benchmarks.} We evaluate chat language models on conversational tasks with human pairwise evaluations and on standard NLP benchmarks with automated metrics, then study whether scores on computationally inexpensive and fast NLP benchmarks are correlated with and predictive of expensive and time-intensive human evaluations.}
    \label{fig:schematic}
\end{figure*}

This shift towards human evaluations raises important questions about the relationship between NLP benchmarks and human evaluations of chat LMs.
Human evaluations have long been considered the gold standard \citep{gatt2018survey,van2019best,celikyilmaz2020evaluation,roller2020opendomainconversationalagentscurrent,van2021human}, but are expensive, time-intensive and noisy, in contrast with computationally cheaper, faster and precise benchmarks.
We explore the relationship between human evaluations and NLP benchmarks in pursuit of understanding what role, if any, benchmarks should play in the era of chat LMs.
We seek to answer two key research questions (Fig.~\ref{fig:schematic}):
\begin{enumerate}
    \item To what extent are human evaluations and NLP benchmarks correlated with one another?
    \item How well can NLP benchmarks predict human evaluations?
\end{enumerate}
To answer these questions, we conducted a large-scale study comparing human evaluations and NLP benchmarks using four Llama 2 Chat language models (LMs) \citep{touvron2023llama2}. For human evaluations, we constructed a large-scale dataset of single-turn and multi-turn prompts from a diverse taxonomy (Fig. \ref{fig:human_eval_prompt_taxonomy}) and collected high quality pairwise human preference data against GPT 3.5 \citep{ouyang2022training} from paid human annotators. For NLP benchmarks, we evaluated the same four Chat Llama 2 models on standard NLP benchmarks under established evaluation processes (metrics, prompting, 0-shot/few-shot, etc.). We analyzed pairwise correlations between NLP benchmark and human evaluations to identify which NLP benchmarks correlate highly with human evaluations and which do not.
We identified which human evaluations, if any, are uncorrelated with any NLP benchmarks.
We then pivoted to predict human evaluations from NLP benchmarks using overparameterized linear regressions and leave-one-out cross-validation, answering the extent to which NLP benchmarks can predict human evaluations. See Appendix~\ref{app:sec:related_work} for Related Work.

\section{Methods: Models, Human Evaluations and NLP Benchmarks}
\label{sec:methods}

We briefly outline our methodology here; for additional information, please see Appendix \ref{app:sec:experimental_methodology}.

\begin{figure*}[t!]
    \centering
    \includegraphics[width=0.9\textwidth]{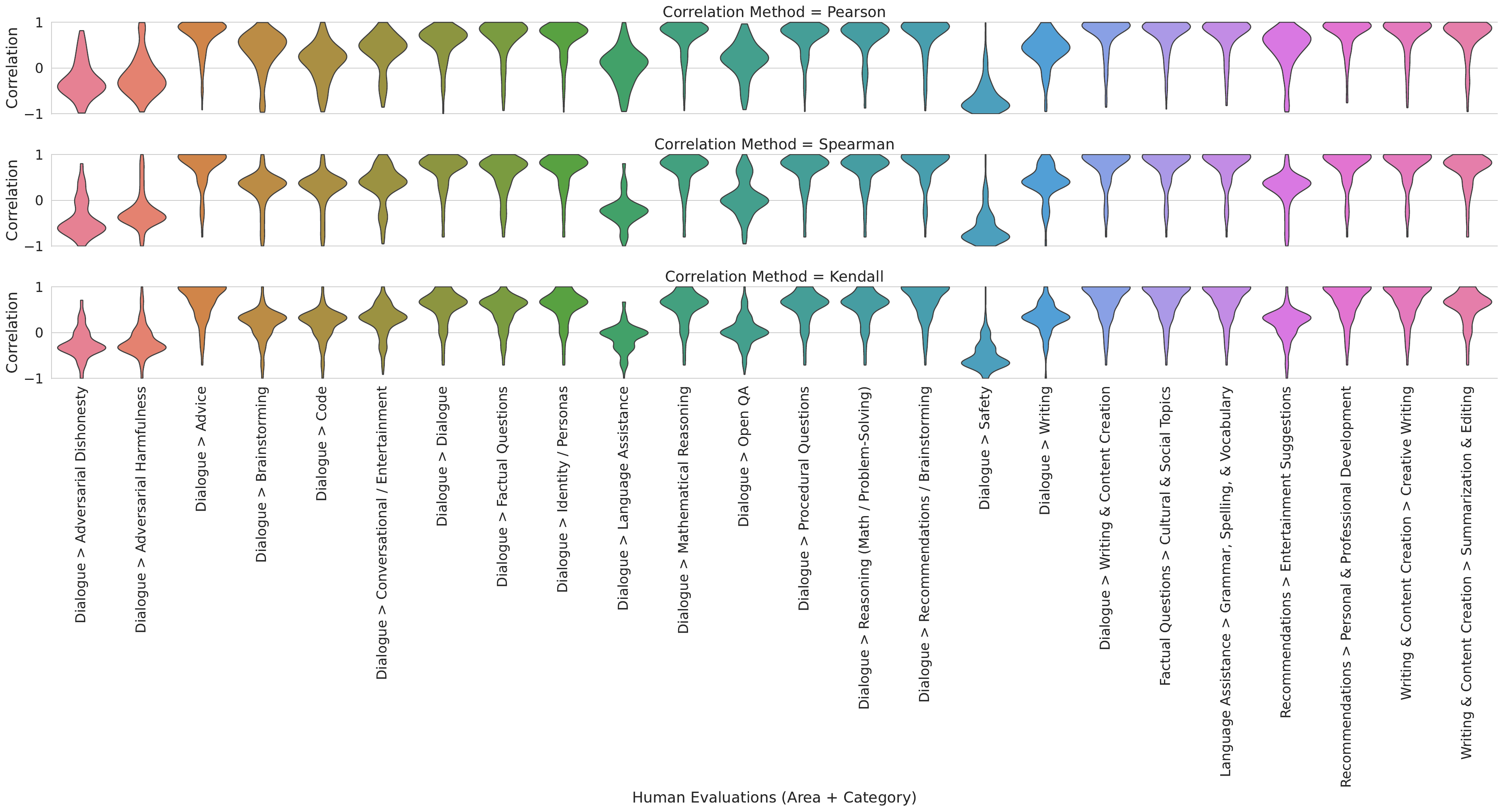}
    \caption{\textbf{Distributions of Correlations between Human Evaluations and NLP benchmarks.} Macroscopically, for each human evaluation area, Chat LM scores are typically highly correlated with NLP benchmarks. Mesoscopically, human and NLP benchmarks remain positively correlated, with notable exceptions: Adversarial Dishonesty, Adversarial Harmfulness and Safety are anticorrelated with most NLP benchmarks, and Language Assistance and Open QA are uncorrelated.}
    \label{fig:corr:human_eval_area}
\end{figure*}

\textbf{Models.} We evaluated four Chat Llama 2 models with 7, 13, 34, and 70 billion parameters pre-trained on 2 trillion tokens and finetuned using supervised finetuning \citep{sanh2021multitask, chung2022scaling, longpre2023flan} and reinforcement learning from human feedback \citep{christiano2017deep, ziegler2019fine, stiennon2020learning}.
We chose the Llama 2 family because, at the time we collected our data, it contained leading open-access chat-finetuned models spanning multiple scales with minimal variations, ensuring a consistent foundation for our investigations.

\textbf{Human Evaluations: Single Turn \& Multi-Turn.} In this work, our aim was specifically to identify which NLP benchmark scores are predictive of human preferences on open-ended prompts representative of real-world chat model usage. We chose this approach to maximize the ecological validity and generalizability of the findings to real-world use cases. For a concrete example, we may want our chat language models (LMs) to excel at providing bespoke career advice; which NLP benchmarks provide useful signals for whether models are improving at such tasks?

To answer such questions, we created a taxonomy of single-turn and multi-turn interactions (Fig. \ref{fig:human_eval_prompt_taxonomy}) between chat LMs and humans. For single-turn interactions, we generated a diverse set of prompts spanning common areas of interest: Factual Questions, Procedural Questions, Language Assistance, Writing \& Content Creation, Dialogue, Code, Reasoning, Recommendations / Brainstorming and Safety, with nested categories and subcategories. 
For multi-turn prompts, non-annotator humans were asked to have conversations (3 to 15 turns long) with all models on similar topics of interest: Factual Questions, Procedural Questions, Language Assistance, Writing \& Content Creation, Summarization \& Editing, General Dialogue, Reasoning and Recommendations / Brainstorming. This taxonomy was chosen to broadly cover common use-cases of Chat LMs.
Example prompts include: ``What is the tallest mountain in the world?" (Factual Question); ``How do I make minestrone soup?" (Procedural Question); ``Please make this sentence more friendly: I need you to stop parking in my space" (Language Assistance). 
See Appendix \ref{app:sec:experimental_methodology:human_evals} for more information.

We paid human annotators to evaluate each of the four Chat Llama 2 models against ChatGPT 3.5 \citep{ouyang2022training} (gpt-3.5-0301) on a dataset of single-turn and multi-turn prompts (Fig \ref{fig:human_eval_prompt_taxonomy}). We chose gpt-3.5-0301 because, at the time this data was collected, gpt-3.5-0301 was a good balance of three desirable properties for our study: performant, cheap, and stable.
For each pair of conversations (one conversation with Chat Llama responses and the other with ChatGPT responses), at least three unique human annotators independently indicated which conversation was preferred using a \citet{likert1932technique} scale from 1 to 7, where 1 denotes the Chat Llama model was strongly preferred and 7 denotes gpt-3.5-0301 was strongly preferred.
Across the 11291 single-turn samples and 2081 multi-turn samples, we had at least 3 unique human annotators evaluate each per pairwise comparison, with 2104 unique annotators overall. 
We averaged the annotators' scores for each pairwise comparison to give us an average human evaluation score per datum.

\textbf{Natural Language Processing (NLP) Benchmarks.} We evaluated the four Chat Llama 2 models on large-scale and commonly-used NLP benchmarks: AGI Eval \citep{zhong2023agieval}, AI2 Reasoning Challenge (ARC; both Easy and Hard) \citep{clark2018arc}, BIG Bench Hard \citep{srivastava2022beyond,suzgun2022challenging} BoolQ \citep{clark2019boolq}, CommonSenseQA \citep{talmor2019commonsenseqa}, COPA \citep{roemmele2011choice}, DROP \citep{dua2019drop}, GSM8K \citep{cobbe2021training}, HellaSwag \citep{zellers2019hellaswag}, HumanEval \citep{chen2021evaluatinglargelanguagemodels}, InverseScaling \citep{mckenzie2022inverse,mckenzie2022round1,mckenzie2022round2}, MBPP \citep{austin2021program}, MMLU \citep{hendrycks2020measuring}, Natural Questions \citep{kwiatkowski2019naturalquestions}, OpenbookQA \citep{mihaylov2018openbookqa}, PIQA \citep{bisk2020piqa}, QuAC \citep{choi2018quac}, RACE \citep{lai2017race}, SIQA \citep{sap2019social}, SQUAD \citep{rajpurkar2016squad}, TLDR \citep{volske2017tl}, TriviaQA \citep{joshi2017triviaqa}, WinoGrande \citep{sakaguchi2021winogrande} and XSum \citep{narayan2018xsum}. Some of these benchmarks (e.g., MMLU) contain subsets (e.g., Jurisprudence) that we do not aggregate over.
These tasks cover commonsense reasoning, world knowledge, reading comprehension, coding and more. We used standard evaluation processes for all academic benchmarks including prompt formatting, metrics, 0-shot/few-shot, etc.
This structured approach facilitates an exhaustive examination of model performances across varied metrics.
For more information, see Appendix \ref{app:sec:experimental_methodology:nlp_benchmarks}.

\begin{figure*}[t!]
    \centering
    \includegraphics[width=\textwidth]{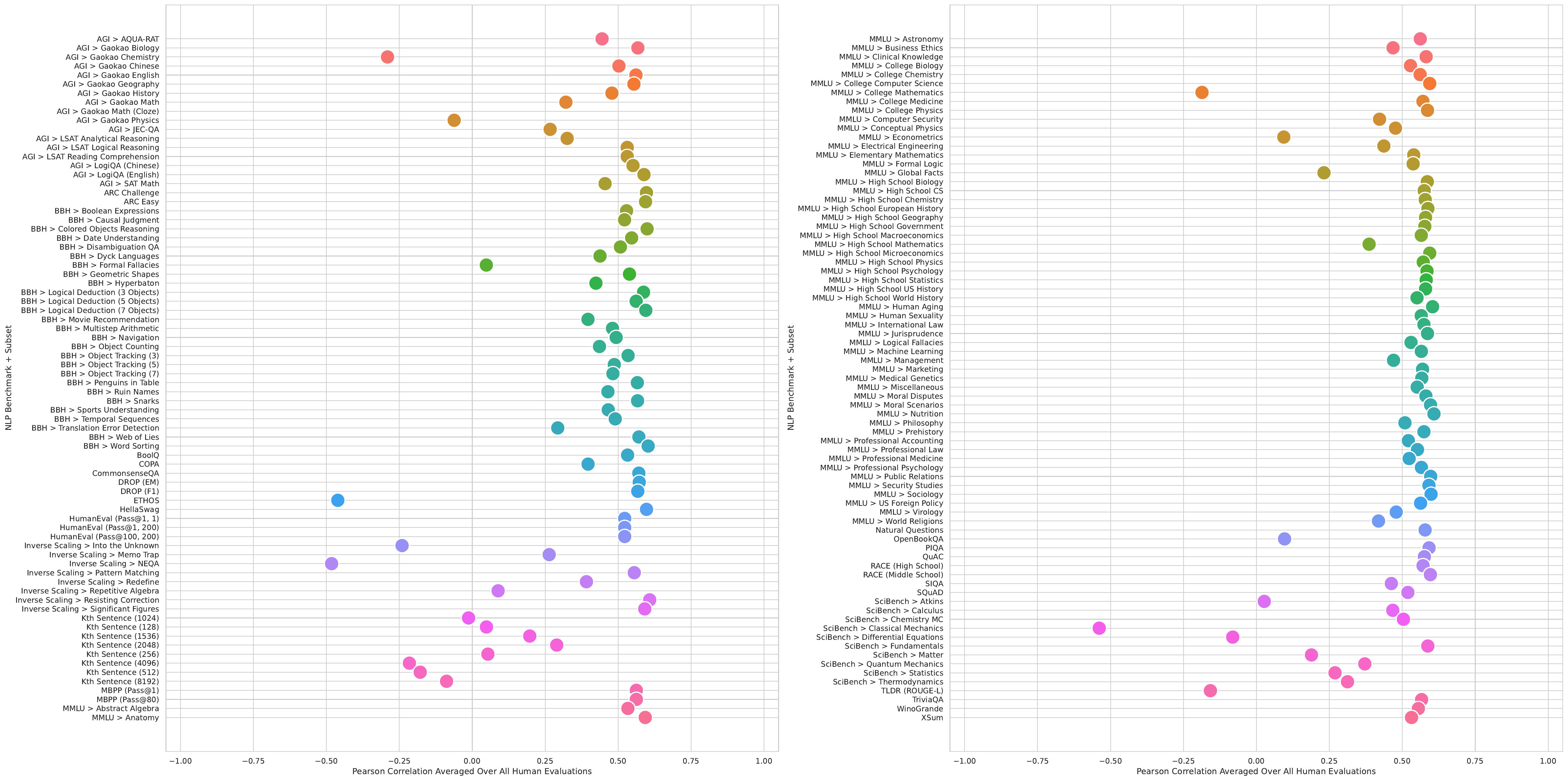}
    \caption{\textbf{NLP Benchmarks Ranked by Average Pearson Correlation over All Human Evaluations.} Certain benchmarks have higher correlations with human evaluations, including a subset of MMLU, a subset of BIG Bench Hard, HellaSwag, ARC, RACE, PIQA, NaturalQuestions, QuAC, and CommonSenseQA. Other benchmarks were weakly or uncorrelated with human evaluations: ETHOS, Kth Sentence, Inverse Scaling (with the exception of Resisting Correction Classification), OpenBookQA, COPA, SciBench and SIQA.}\label{fig:corr:academic_vs_correlation_split_corrmethod}
\end{figure*}

\textbf{Scores for Subsequent Analyses.} For each dataset and evaluation process, we average each model's scores across all samples, yielding two matrices of scores:
$$X_{\text{NLP}} \in \mathbb{R}^{160 \times 4} \quad \quad \quad \quad X_{\text{Human}} \in \mathbb{R}^{55 \times 4}$$
Here, $4$ is the number of models, $160$ is the number of NLP benchmarks per model and $55$ is the number of human evaluation area-category-subcategory scores per model. We subsequently study the correlations between $X_{\text{NLP}}$ and $X_{\text{Human}}$, then test how well $X_{\text{NLP}}$ can predict $X_{\text{Human}}$.

\section{Correlating Human Evaluations with NLP Benchmarks}
\label{sec:correlations}

\begin{figure*}[t!]
    \centering
    \includegraphics[width=\linewidth]{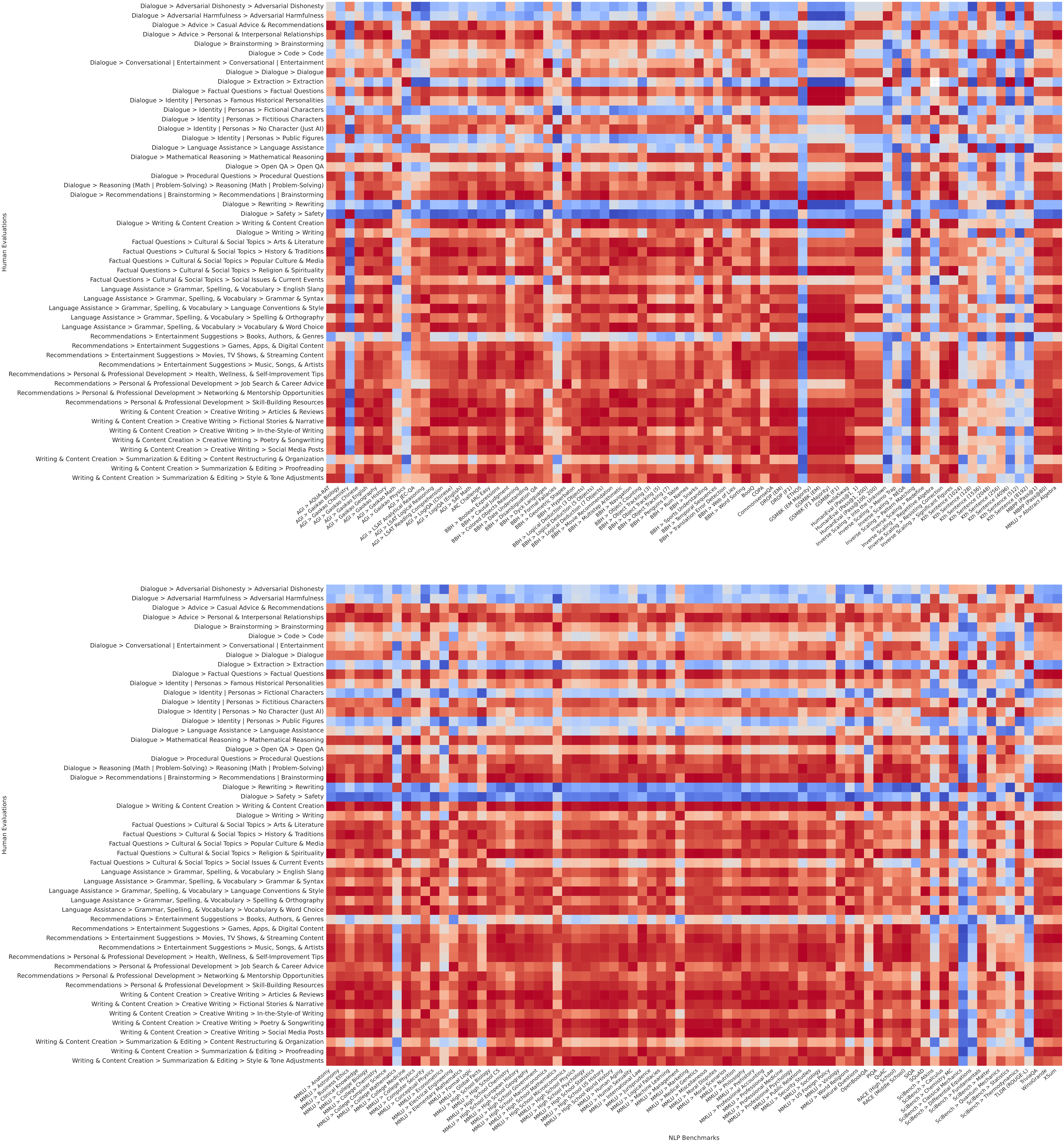}
    \caption{\textbf{Pearson Correlations Between Human Evaluations and NLP Benchmarks.} Rows: Human evaluation areas-categories-subcategories. Columns: NLP benchmarks. The heatmap is row-wrapped to fit on the page. \textcolor{red}{Large positive correlations (+1) are shown in red.} \textcolor{blue}{Large negative anticorrelations (-1) are shown in blue.} Low uncorrelations ($\sim$0) are shown in light-white-gray.}
    \label{fig:corr:microscopic}
\end{figure*}

We began by computing correlations between human evaluations and NLP benchmarks over the 4 average scores per model with three standard correlations  --- Pearson \citep{galton1877heredity}, Spearman \citep{spearman1904proof} and Kendall \citep{kendall1938new} --- giving us three correlation matrices of shape $160 \times 55$ between every pair of NLP benchmark and human evaluation area-category-subcategory (Fig. \ref{fig:corr:microscopic}).
By using different correlation metrics, we aim to robustly characterize the relationships between human and NLP benchmarks.

Macroscopically, at the most coarse grouping of human evaluations in our taxonomy (i.e., areas) (Fig.~\ref{fig:human_eval_prompt_taxonomy}), we found that average NLP benchmark scores are highly correlated with average human scores for all human evaluation areas under all three correlation metrics (Fig. \ref{fig:corr:human_eval_area} top).
These strong correlations suggest that, at a high level, NLP benchmarks are reasonable proxies for human judgments of LM quality.

Mesoscopically, at the level of human evaluation areas and categories, NLP benchmarks remain highly correlated with human evaluations, with two notable types of exceptions (Fig. \ref{fig:corr:human_eval_area}). First, Adversarial Dishonesty, Adversarial Harmfulness, and Safety are anti-correlated with most NLP benchmarks, potentially indicating that these adversarial and safety-focused categories are more easily transgressed by more capable LMs; an alternative hypothesis could be that safety benchmarks simply are not especially good, as demonstrated by \citet{ren2024safetywashing}. Second, Language Assistance and Open Question Answering are uncorrelated with most NLP benchmarks, suggesting that these categories may require new NLP benchmarks. Open Question Answering was surprising given that some of our NLP benchmarks are open question answering datasets, e.g., OpenBookQA \citep{mihaylov2018openbookqa}. We found the three correlations metrics visually agreed with one another and were themselves tightly coupled (App. Fig. \ref{app:fig:correlation_couplings}), and so we present only one (Pearson) moving forward, with equivalent plots of the other two (Spearman, Kendall) deferred to the appendix.

\subsection{Which human evaluations have few-to-no correlated NLP benchmarks?}
\label{sec:correlations:subsec:nlp_benchmarks_no_correlations}

To the best of our ability to discern, none. Every human evaluation seemed to have at least some NLP benchmarks that were either correlated or anticorrelated with it. This result is promising because it suggests human evaluations might be predictable from NLP benchmarks (Sec.~\ref{sec:predictions}).  

\subsection{Which NLP benchmarks exhibit high correlations with human evaluations?}
\label{sec:correlations:subsec:nlp_benchmarks_highest_correlations}

\begin{figure*}[t!]
    \centering
    \includegraphics[width=0.9\textwidth]{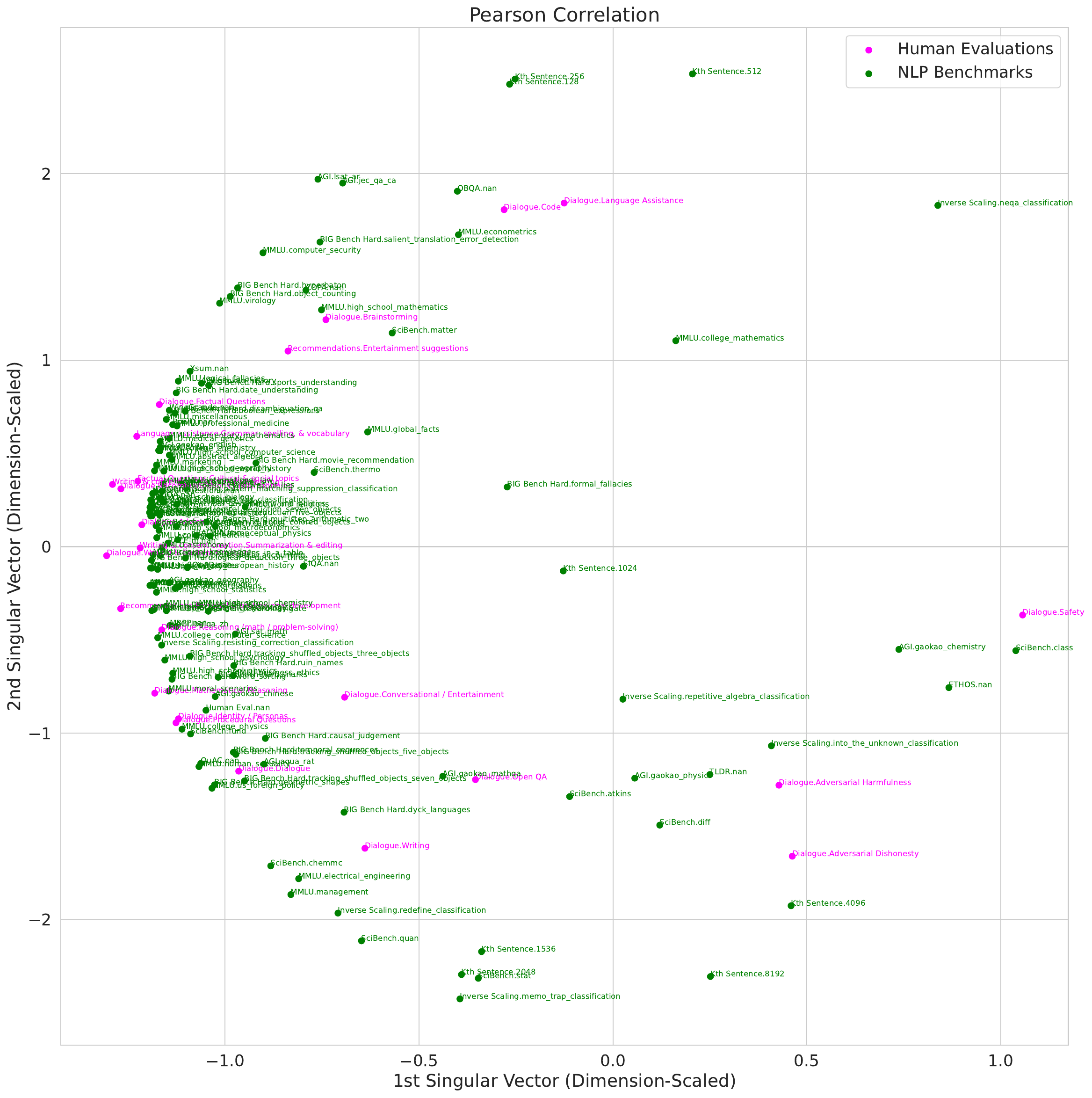}
    \caption{\textbf{Matrix Decomposition of Pairwise Pearson Correlations Between Human Evaluations and NLP Benchmarks.} The correlation matrix has 3 non-zero singular values (App. Fig. \ref{app:fig:academic_human_singular_value_spectra}). Bottom: \textcolor{Magenta}{Human evaluations} and \textcolor{Green}{NLP benchmarks} are plotted projected along the (dimension-scaled) first two singular modes of the Pearson correlation matrix. The bulk of evaluations live in one community (left), with smaller communities (top, bottom, right); for an in-depth interpretation, see Sec. \ref{sec:correlations:subsec:community_detection}.}
    \label{fig:corr:singular_modes}
\end{figure*}

To answer this question, we ordered NLP benchmarks based on their average correlation score with all human evaluation areas, categories and subcategories.
We found many NLP benchmarks have high average correlation with human evaluations (Fig. \ref{fig:corr:academic_vs_correlation_split_corrmethod}); the highest average correlation NLP benchmarks include a subset of MMLU (Nutrition, Human Aging, Sociology, Public Relations, Moral Scenarios, College Computer Science), a subset of BIG Bench Hard (Word Sorting, Reasoning About Colored Objects, Logical Deduction), HellaSwag, ARC, RACE, PIQA, NaturalQuestions, QuAC, CommonSenseQA, DROP and TriviaQA. Other benchmarks were less correlated or uncorrelated with human evaluations: ETHOS, Kth Sentence, Inverse Scaling (excluding Resisting Correction Classification), OpenBookQA, COPA, SciBench (excluding Fundamentals of Physics) and SIQA.
Upon investigating, some of the most highly correlated NLP benchmarks make sense. For instance, Inverse Scaling's Resisting Correction Classification ranked second highest for being correlated with human evaluations, and the task measures a highly desirable capability for human users: the LM's ability to follow user instructions that run counter to the LM's natural inclinations.

\subsection{Matrix Decomposition of Correlations between Human Evaluations and NLP Benchmarks} \label{sec:correlations:subsec:community_detection}

To gain a structured view of how human evaluations and NLP benchmarks interrelate, we analyze the $160 \times 55$ Pearson correlation matrix between them by computing the singular value decomposition of the correlation matrix $C = U \Sigma V^T$, where $U \in \mathbb{R}^{160 \times 3}$ and $V \in \mathbb{R}^{55 \times 3}$ are the left and right singular vectors. 
We observe only three non-zero singular values (Appendix Figure~\ref{app:fig:academic_human_singular_value_spectra}), indicating that most of the variance in correlations is captured by three underlying dimensions.
Interpreting each row of $U$ and $V$ as coordinates in the corresponding low-dimensional space provides a way to visualize how tasks and human evaluations group together based on their similarity in correlation patterns, as shown by plotting each NLP benchmark (\textcolor{Green}{green}) and each human evaluation category (\textcolor{Magenta}{magenta}) in the 2D space spanned by the first two singular vectors (scaled by their singular values). (Fig.~\ref{fig:corr:singular_modes}).

\textbf{Overall Alignment (Left Cluster).} The largest group of points, combining both \textcolor{Magenta}{human evaluations} and \textcolor{Green}{benchmarks}, sits on the left side. This indicates that many standard NLP benchmarks (e.g., language understanding, commonsense reasoning, factual QA) tend to move in tandem with broad human-judged performance (e.g., correctness, clarity) across our four models.

\textbf{Dialogue and Context-Oriented Tasks (Top-Left).} Certain dialogue-related human evaluations (e.g., \textcolor{Magenta}{Language Assistance, Coding Assistance}) appear near several benchmarks that emphasize context or discourse (e.g., \textcolor{Green}{OpenBookQA}, \textcolor{Green}{Kth Sentence}). Their proximity suggests a shared correlation pattern across the models, possibly reflecting reliance on social reasoning and contextual cues.

\textbf{Adversarial and Safety-Focused Tasks (Right \& Bottom-Right).} Evaluations tied to \textcolor{Magenta}{Adversarial Harmfulness/Dishonesty} and \textcolor{Magenta}{Safety} show distinct positions, often near benchmarks aimed at exposing errors or biases (e.g., \textcolor{Green}{Inverse Scaling tasks}, \textcolor{Green}{ETHOS} for hate speech). This segregation indicates that safety/adversarial capabilities differ in how they correlate with more conventional tasks.

\textbf{Open QA \& Domain Knowledge (Lower-Left).} Finally, \textcolor{Magenta}{Open QA} and some \textcolor{Magenta}{Writing} evaluations lie closer to benchmarks demanding specialized knowledge (\textcolor{Green}{MMLU.Electrical Engineering}, \textcolor{Green}{SciBench.Quantum Chemistry}), suggesting that open-ended user queries may align more with advanced domain-knowledge benchmarks than with simpler tasks.

Overall, this matrix decomposition shows that while there is a dominant “general ability” factor (represented by the first singular value) that aligns most tasks and evaluations, additional singular vectors capture subtler patterns. These include the safety/adversarial dimension and context-dependent or domain-specialized dimensions. Consequently, the correlations between human evaluations and NLP benchmarks exhibit a rich low-rank structure indicative of multiple underlying performance factors.

\begin{figure*}[t!]
    \centering
    \includegraphics[width=0.9\textwidth]{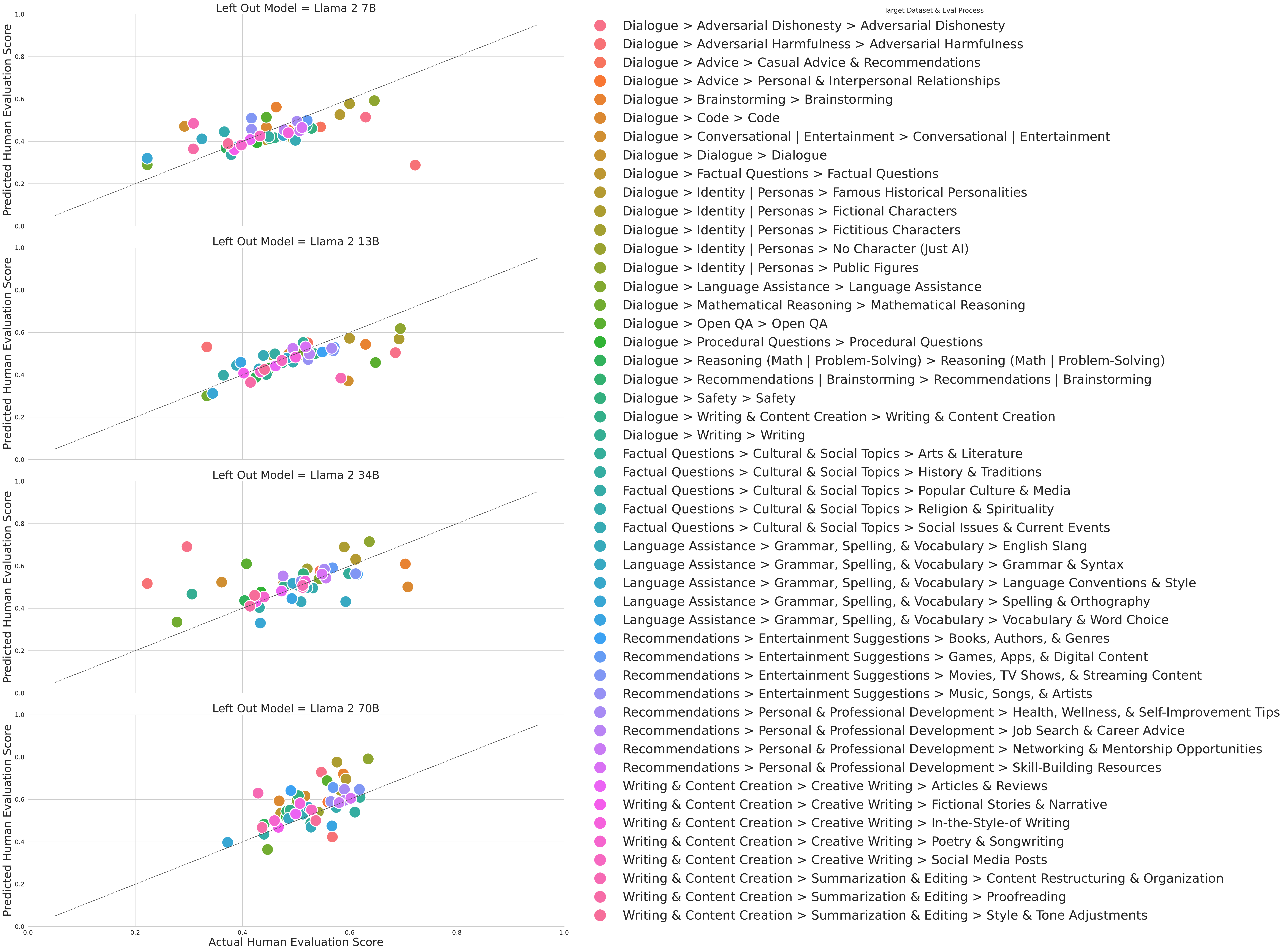}
    \caption{\textbf{Leave-One-Out Cross Validated Linear Regression Predictions of Human Evaluations.} Linear regressions accurately predict human evaluation scores from all NLP benchmark scores. Each subfigure shows predicted human evaluation scores against actual human evaluation scores on each of the four left-out Chat Llama 2 models colored by human evaluation area, category and subcategory.}
    \label{fig:reg:overparameterized_regressions_leave_one_out}
\end{figure*}
\section{Predicting Human Evaluations from NLP Benchmarks}
\label{sec:predictions}

Having established the existence of correlations between human evaluations and NLP benchmarks, we investigated the feasibility of predicting human evaluations from NLP benchmarks. Our goal is to build predictive models that accurately predict a language model's average human evaluation scores per areas and categories using the model's average scores on NLP benchmarks and tasks. However, we faced a significant challenge due to the overparameterized nature of our data: for each target human evaluation area or category, there are approximately 150 covariates (NLP benchmarks and tasks) but only four models.

\textbf{Predictive Modeling: Overparameterized Linear Regressions.}
To predict human evaluations from NLP benchmarks, we used overparameterized linear regression. In general, overparameterized linear regression is known to be capable of generalizing (App. Sec. \ref{app:sec:generalization_of_overparameterized_models}), although whether linear models would generalize in this setting was an empirical question. 
For each human evaluation area and category, we fit a linear model to predict a language model's average human evaluation score from its average scores on all NLP benchmarks and tasks. To assess the predictive accuracy of these overparameterized models, we employed leave-one-out cross validation: we fit four separate linear models, each time fitting on three of the chat LMS' scores and holding out the fourth to test the performance of the linear model. This approach allows us to estimate the models' performance on unseen data, albeit with limitations due to the small sample size.
Before fitting the models, we normalized all human evaluation scores to lie in $[0, 1]$ rather than $[-7, -1]$ (recalling that higher scores indicate the human evaluator prefers the Chat Llama 2 model compared to GPT-3.5). 

\textbf{Results.} Across human evaluation areas and categories, we found that the linear models' predicted average human evaluation scores generally align well with the actual average human evaluation scores, as evidenced by most points falling close to the identity line in the predicted score vs. actual score plane (Fig. \ref{fig:reg:overparameterized_regressions_leave_one_out}). This suggests that, despite the overparameterization, the linear models can capture meaningful relationships between NLP benchmarks and human evaluations. However, we caution against over-interpreting these results, as the small sample size and the assumption of linearity may limit the generalizability of these findings to other language models or evaluation settings.

To gain insight into which NLP benchmarks are most informative for predicting human evaluation scores, we examine the learned weights of the linear models (Fig. \ref{app:fig:linear_regression_coefficients}). NLP benchmarks with consistently high absolute weights across different human evaluation areas and categories are likely to be more predictive of human judgments. However, due to the overparameterized nature of the models, we refrain from drawing strong conclusions about the relative importance of individual benchmarks and instead focus on the overall predictive performance.
These results suggest that scaling up the number of chat LMs and human evaluation data could unlock highly predictive models of slow, noisy and expensive but valuable human evaluations using fast, precise and cheaper NLP benchmarks.
\section{Discussion}

In this paper, we explored the relationship between human evaluations and NLP benchmarks of chat-finetuned language models (chat LMs). Our work is motivated by the recent shift towards human evaluations as the primary means of assessing chat LM performance, and the desire to determine the role that NLP benchmarks should play.

Through a large-scale study of the Chat Llama 2 model family on a diverse set of human and NLP evaluations, we demonstrated that NLP benchmarks are generally well-correlated with human judgments of chat LM quality. However, our analysis also reveals some notable exceptions to this overall trend. In particular, we find that adversarial and safety-focused evaluations, as well as language assistance and open question answering tasks, exhibit weaker or negative correlations respectively with NLP benchmarks. We also explored predicting human evaluation scores from NLP evaluation scores using overparameterized linear regression models. Our results suggest that NLP benchmarks can indeed be used to predict aggregate human preferences, although we caution that the limited sample size and the assumptions of our models may limit the generalizability of these findings. Our results suggest that NLP benchmarks can serve as fast and cheap proxies of slower and expensive human evaluations in assessing chat LMs.

Additionally, our work highlights the need for further research into NLP evaluations that can effectively capture important aspects of LM behavior, such as safety, robustness to adversarial inputs, and performance on complex, open-ended tasks. It is possible that new NLP benchmarks can provide signals on these topics, e.g., \citep{wang2023decodingtrust}. Of particular interest is developing human-interpretable and scaling-predictable evaluation processes, e.g., \citep{schaeffer2024emergent, ruan2024observational,schaeffer2024predictingdownstreamcapabilitiesfrontier}. Developing and refining such evaluation methods \citep{madaan2024quantifyingvarianceevaluationbenchmarks}, as well as detecting whether evaluations scores faithfully capture models' true performance \citep{oren2023proving,schaeffer2023pretrainingtestsetneed,roberts2023cutoff,jiang2024investigatingdatacontaminationpretraining,zhang2024careful,duan2024uncoveringlatentmemoriesassessing} will be crucial for ensuring that LMs are safe, reliable, and beneficial as they become increasingly integrated into society.


\clearpage

\bibliography{references_rylan}
\bibliographystyle{icml2025}

\clearpage
\appendix
\onecolumn

\clearpage

\section{Related Work}
\label{app:sec:related_work}

The evaluation of language models has a rich and constantly evolving history. Human evaluations have long been considered the gold standard \citep{gatt2018survey,van2019best,celikyilmaz2020evaluation,roller2020opendomainconversationalagentscurrent,van2021human}, despite serious objections raised regarding the collection, analysis, and interpretation of human evaluation scores \citep{novikova2018rankme,howcroft2020twenty, bowman2021fixbenchmarkingnaturallanguage,karpinska2021perilsusingmechanicalturk, clark2021all,smith2022humanevaluationconversationsopen, gehrmann2023repairing,finch2023dontforgetabcsevaluating}. Many classic NLP benchmark metrics, such as BLEU \citep{papineni2002bleu}, NIST \citep{doddington2002nist}, ROUGE \citep{lin2004rouge}, and METEOR \citep{banerjee2005meteor}, were introduced on the premise that they correlate with human judgments. However, subsequent studies revealed that the relationship between automated metrics and human evaluations is often complex and not straightforward \citep{liu2016hownot, novikova2017we, reiter2018structured, karpinska2021perilsusingmechanicalturk}. Another prominent class of evaluation methods are based on machine learning models, e.g., word mover distance \citep{kusner2015word} and BERT-Score \citep{zhang2019bertscore} that have since evolved into using chat LMs themselves as evaluators \citep{wang2023chatgpt,zheng2024judging, chiang2023largelanguagemodelsalternative,chan2023chatevalbetterllmbasedevaluators,bavaresco2024llms,fu2024gptscore}, albeit with limitations, e.g., \citep{dorner2024limits,szymanski2024limitationsllmasajudgeapproachevaluating, thakur2024judging}.

The earliest investigations into the general relationship between NLP benchmark scores and human evaluations date back to \citet{bangalore2000evaluation}, \citet{belz2006comparing}, and \citet{liu2016hownot}. In the context of natural language generation, \citet{clinciu2021study} found that embedding-based automated metrics (e.g., BERT-Score \citep{zhang2019bertscore} and BLEURT \citet{sellam2020bleurt}) correlate more strongly with human judgments compared to word-overlap metrics (e.g., ROUGE \citep{lin2004rouge} and BLEU \citep{papineni2002bleu}). In the domain of natural language inference, \citet{schuff2021does} found that automated metrics do not appear to correlate with human judgment scores. However, the majority of these works predate the current era of chat LMs, which exhibit significantly more advanced capabilities compared to their predecessors. This new era motivates our work to investigate the relationship between NLP benchmarks and human evaluations when evaluating chat LMs.

\section{Experimental Methodology: Data and Analyses}
\label{app:sec:experimental_methodology}

\subsection{Data: Natural Language Processing (NLP) Benchmark Scores}
\label{app:sec:experimental_methodology:nlp_benchmarks}

We chose which NLP benchmarks to include based largely on which frontier AI models were reporting performance scores on. Llama 1 \citep{touvron2023llama1} and Llama 2 \citep{touvron2023llama2} were our primary guides, as were Gemini 1 and Gemini 1.5 \citep{team2023gemini, reid2024gemini}, Claude 3 \citep{anthropic2023claude3}, and Mistral \citep{jiang2023mistral}.
We evaluated the 4 Llama 2 Chat models following the same evaluation processes reported in the Llama 2 paper \citep{touvron2023llama2}.
This included matching the prompt formatting, automated metric scoring and (when appropriate) few-shot prompting and chain-of-thought prompting.
After evaluating the four Chat Llama 2 models on these NLP benchmarks and evaluation processes, we obtained 160 scores per model for our analyses.

\begin{figure*}[t!]
    \centering
    \includegraphics[width=0.9\textwidth]{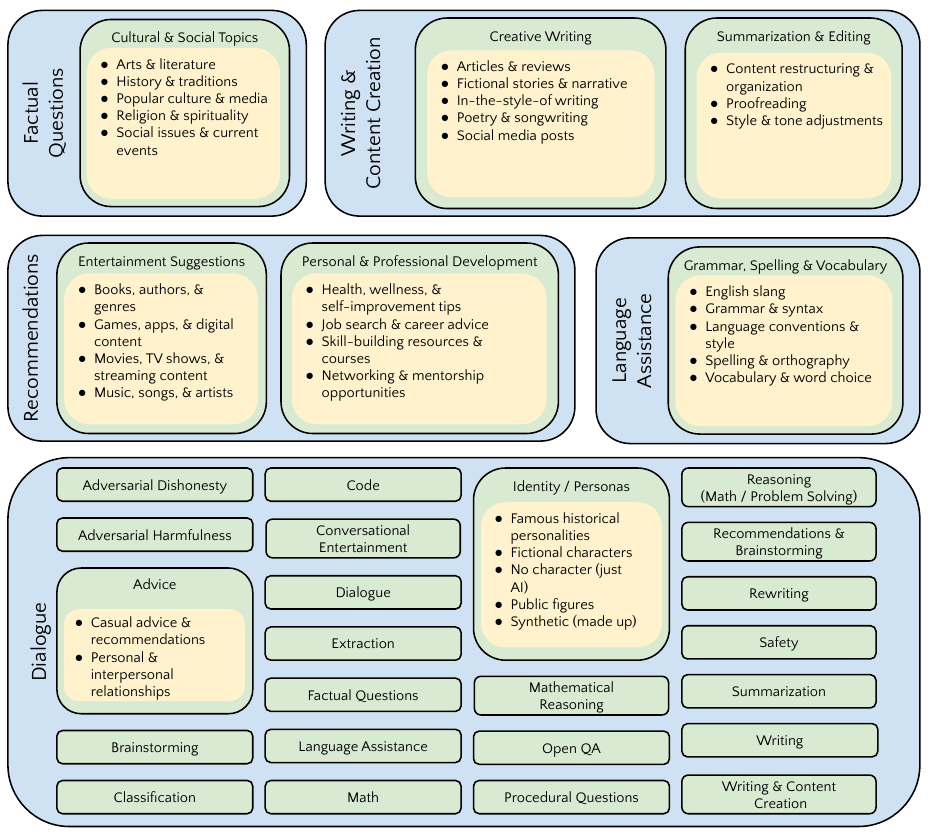}
    \caption{\textbf{Human Evaluations: Taxonomy of Single-Turn and Multi-Turn Conversations.} Single-turn and multi-turn prompts were created in a hierarchical taxonomy of 9 areas (blue), categories (green) and subcategories (yellow). Paid human annotators then rated Chat Llama 2 generations against ChatGPT generations by paid human annotators on a 7 point Likert scale \citep{likert1932technique}.}
    \label{fig:human_eval_prompt_taxonomy}
\end{figure*}

\begin{longtable}{|c|c|c|c|c|}
    \caption{\textbf{Natural Language Processing Datasets and Evaluation Processes.} In this work, we used the above well-established NLP datasets and evaluation processes. ``cot" means Chain-of-Thought prompting \citep{cobbe2021training, wei2022chain}. ``Gen" refers generations per evaluation. For more information, please see Section \ref{sec:methods} and Appendix \ref{app:sec:experimental_methodology}.}
    \label{app:tab:all_benchmarks_tasks}\\
    \hline
    Benchmark & Subset & Metric & Shots & Additional\\
    \hline
    \endfirsthead
    \multicolumn{5}{c}%
    {{\bfseries \tablename\ \thetable{} -- continued from previous page}} \\
    \hline
    Benchmark & Subset & Metric & Shots & Additional \\
    \hline
    \endhead
    AGI & aqua\_rat & Acc Char & 5 & -- \\
    AGI & gaokao\_biology & Acc Char & 5 & -- \\
    AGI & gaokao\_chemistry & Acc Char & 5 & -- \\
    AGI & gaokao\_chinese & Acc Char & 5 & -- \\
    AGI & gaokao\_english & Acc Char & 5 & -- \\
    AGI & gaokao\_geography & Acc Char & 5 & -- \\
    AGI & gaokao\_history & Acc Char & 5 & -- \\
    AGI & gaokao\_mathcloze & Exact Match & 5 & -- \\
    AGI & gaokao\_mathqa & Acc Char & 5 & -- \\
    AGI & gaokao\_physics & Acc Char & 5 & -- \\
    AGI & jec\_qa\_ca & Acc Char & 5 & -- \\
    AGI & jec\_qa\_kd & Acc Char & 5 & -- \\
    AGI & logiqa\_en & Acc Char & 3 & -- \\
    AGI & logiqa\_zh & Acc Char & 3 & -- \\
    AGI & lsat\_ar & Acc Char & 3 & -- \\
    AGI & lsat\_lr & Acc Char & 3 & -- \\
    AGI & lsat\_rc & Acc Char & 3 & -- \\
    AGI & sat\_math & Acc Char & 5 & -- \\
    ARC & Challenge & Acc Char & 0 & -- \\
    ARC & Easy & Acc Char & 0 & -- \\
    BBH & boolean\_expressions & Exact Match & 0 & -- \\
    BBH & causal\_judgement & Exact Match & 0 & -- \\
    BBH & date\_understanding & Exact Match & 0 & -- \\
    BBH & disambiguation\_qa & Exact Match & 0 & -- \\
    BBH & dyck\_languages & Exact Match & 0 & -- \\
    BBH & formal\_fallacies & Exact Match & 0 & -- \\
    BBH & geometric\_shapes & Exact Match & 0 & -- \\
    BBH & hyperbaton & Exact Match & 0 & -- \\
    BBH & logical\_deduction\_3\_objects & Exact Match & 0 & -- \\
    BBH & logical\_deduction\_5\_objects & Exact Match & 0 & -- \\
    BBH & logical\_deduction\_7\_objects & Exact Match & 0 & -- \\
    BBH & movie\_recommendation & Exact Match & 0 & -- \\
    BBH & multistep\_arithmetic\_two & Exact Match & 0 & -- \\
    BBH & navigate & Exact Match & 0 & -- \\
    BBH & object\_counting & Exact Match & 0 & -- \\
    BBH & penguins\_in\_a\_table & Exact Match & 0 & -- \\
    BBH & reasoning\_about\_colored\_objects & Exact Match & 0 & -- \\
    BBH & ruin\_names & Exact Match & 0 & -- \\
    BBH & salient\_translation\_error\_detection & Exact Match & 0 & -- \\
    BBH & snarks & Exact Match & 0 & -- \\
    BBH & sports\_understanding & Exact Match & 0 & -- \\
    BBH & temporal\_sequences & Exact Match & 0 & -- \\
    BBH & tracking\_shuffled\_objects\_3 & Exact Match & 0 & -- \\
    BBH & tracking\_shuffled\_objects\_5 & Exact Match & 0 & -- \\
    BBH & tracking\_shuffled\_objects\_7 & Exact Match & 0 & -- \\
    BBH & web\_of\_lies & Exact Match & 0 & -- \\
    BBH & word\_sorting & Exact Match & 0 & -- \\
    BoolQ & -- & Acc Token & 0 & -- \\
    CommonSenseQA & -- & Acc Char & 7 & -- \\
    COPA & -- & Acc Char & ? & -- \\
    DROP & -- & Exact Match & 0 & -- \\
    DROP & -- & F1 & 0 & -- \\
    ETHOS & -- & Acc Char & 0 & -- \\
    GSM8K & -- & Exact Match & 0 & 100 Gen \\
    GSM8K & -- & F1 & 0 & 100 Gen \\
    GSM8K & -- & Exct Mtch maj1@100 & 0 & 100 Gen \\
    GSM8K & -- & F1 maj1@100 & 0 & 100 Gen \\
    HellaSwag & -- & Acc Char & ? & -- \\
    Human Eval & -- & pass@1 & 0 & 1 Gen \\
    Human Eval & -- & pass@1 & 0 & 200 Gen \\
    Human Eval & -- & pass@100 & 0 & 200 Gen \\
    Inverse Scaling & hindsight\_neglect & Exact Match & 0 & -- \\
    Inverse Scaling & into\_the\_unknown & Exact Match & 0 & -- \\
    Inverse Scaling & memo\_trap & Exact Match & 0 & -- \\
    Inverse Scaling & modus\_tollens & Exact Match & 0 & -- \\
    Inverse Scaling & neqa & Exact Match & 0 & -- \\
    Inverse Scaling & pattern\_matching\_suppression & Exact Match & 0 & -- \\
    Inverse Scaling & redefine & Exact Match & 0 & -- \\
    Inverse Scaling & repetitive\_algebra & Exact Match & 0 & -- \\
    Inverse Scaling & resisting\_correction & Exact Match & 0 & -- \\
    Inverse Scaling & sig\_figs & Exact Match & 0 & -- \\
    Kth Sentence & 128 & ROUGE-2 & 0 & -- \\
    Kth Sentence & 256 & ROUGE-2 & 0 & -- \\
    Kth Sentence & 512 & ROUGE-2 & 0 & -- \\
    Kth Sentence & 1024 & ROUGE-2 & 0 & -- \\
    Kth Sentence & 1536 & ROUGE-2 & 0 & -- \\
    Kth Sentence & 2048 & ROUGE-2 & 0 & -- \\
    Kth Sentence & 4096 & ROUGE-2 & 0 & -- \\
    Kth Sentence & 8192 & ROUGE-2 & 0 & -- \\
    MBPP & -- & pass@1 & 3 & 80 Gen \\
    MBPP & -- & pass@80 & 3 & 80 Gen \\
    MMLU & abstract\_algebra & Acc Char & 5 & -- \\
    MMLU & anatomy & Acc Char & 5 & -- \\
    MMLU & astronomy & Acc Char & 5 & -- \\
    MMLU & business\_ethics & Acc Char & 5 & -- \\
    MMLU & clinical\_knowledge & Acc Char & 5 & -- \\
    MMLU & college\_biology & Acc Char & 5 & -- \\
    MMLU & college\_chemistry & Acc Char & 5 & -- \\
    MMLU & college\_computer\_science & Acc Char & 5 & -- \\
    MMLU & college\_mathematics & Acc Char & 5 & -- \\
    MMLU & college\_medicine & Acc Char & 5 & -- \\
    MMLU & college\_physics & Acc Char & 5 & -- \\
    MMLU & computer\_security & Acc Char & 5 & -- \\
    MMLU & conceptual\_physics & Acc Char & 5 & -- \\
    MMLU & econometrics & Acc Char & 5 & -- \\
    MMLU & electrical\_engineering & Acc Char & 5 & -- \\
    MMLU & elementary\_mathematics & Acc Char & 5 & -- \\
    MMLU & formal\_logic & Acc Char & 5 & -- \\
    MMLU & global\_facts & Acc Char & 5 & -- \\
    MMLU & high\_school\_biology & Acc Char & 5 & -- \\
    MMLU & high\_school\_chemistry & Acc Char & 5 & -- \\
    MMLU & high\_school\_computer\_science & Acc Char & 5 & -- \\
    MMLU & high\_school\_european\_history & Acc Char & 5 & -- \\
    MMLU & high\_school\_geography & Acc Char & 5 & -- \\
    MMLU & high\_school\_government\_and\_politics & Acc Char & 5 & -- \\
    MMLU & high\_school\_macroeconomics & Acc Char & 5 & -- \\
    MMLU & high\_school\_mathematics & Acc Char & 5 & -- \\
    MMLU & high\_school\_microeconomics & Acc Char & 5 & -- \\
    MMLU & high\_school\_physics & Acc Char & 5 & -- \\
    MMLU & high\_school\_psychology & Acc Char & 5 & -- \\
    MMLU & high\_school\_statistics & Acc Char & 5 & -- \\
    MMLU & high\_school\_us\_history & Acc Char & 5 & -- \\
    MMLU & high\_school\_world\_history & Acc Char & 5 & -- \\
    MMLU & human\_aging & Acc Char & 5 & -- \\
    MMLU & human\_sexuality & Acc Char & 5 & -- \\
    MMLU & international\_law & Acc Char & 5 & -- \\
    MMLU & jurisprudence & Acc Char & 5 & -- \\
    MMLU & logical\_fallacies & Acc Char & 5 & -- \\
    MMLU & machine\_learning & Acc Char & 5 & -- \\
    MMLU & management & Acc Char & 5 & -- \\
    MMLU & marketing & Acc Char & 5 & -- \\
    MMLU & medical Genetics & Acc Char & 5 & -- \\
    MMLU & miscellaneous & Acc Char & 5 & -- \\
    MMLU & moral\_disputes & Acc Char & 5 & -- \\
    MMLU & moral\_scenarios & Acc Char & 5 & -- \\
    MMLU & nutrition & Acc Char & 5 & -- \\
    MMLU & philosophy & Acc Char & 5 & -- \\
    MMLU & prehistory & Acc Char & 5 & -- \\
    MMLU & professional\_accounting & Acc Char & 5 & -- \\
    MMLU & professional\_law & Acc Char & 5 & -- \\
    MMLU & professional\_medicine & Acc Char & 5 & -- \\
    MMLU & professional\_psychology & Acc Char & 5 & -- \\
    MMLU & public\_relations & Acc Char & 5 & -- \\
    MMLU & security\_studies & Acc Char & 5 & -- \\
    MMLU & sociology & Acc Char & 5 & -- \\
    MMLU & us\_foreign\_policy & Acc Char & 5 & -- \\
    MMLU & virology & Acc Char & 5 & -- \\
    MMLU & world\_religions & Acc Char & 5 & -- \\
    NaturalQuestions & -- & Exact Match & 0 & -- \\
    OpenBookQA & -- & Acc Completion & 0 & -- \\
    PIQA & -- & Acc Char & 0 & -- \\
    QuAC & -- & F1 & 0 & -- \\
    RACE & High School & Acc Char & 0 & -- \\
    RACE & Middle School & Acc Char & 0 & -- \\
    SIQA & -- & Acc Char & 0 & -- \\
    SQuAD & -- & Exact Match & 0 & -- \\
    SciBench & atkins & Fuzzy Match & 0 & -- \\
    SciBench & calculus & Fuzzy Match & 0 & -- \\
    SciBench & chemmc & Fuzzy Match & 0 & -- \\
    SciBench & class & Fuzzy Match & 0 & -- \\
    SciBench & diff & Fuzzy Match & 0 & -- \\
    SciBench & fund & Fuzzy Match & 0 & -- \\
    SciBench & matter & Fuzzy Match & 0 & -- \\
    SciBench & quan & Fuzzy Match & 0 & -- \\
    SciBench & stat & Fuzzy Match & 0 & -- \\
    SciBench & thermo & Fuzzy Match & 0 & -- \\
    TLDR & -- & ROUGE-2 & 0 & -- \\
    TLDR & -- & ROUGE-L & 0 & -- \\
    TriviaQA & -- & Exact Match & 0 & -- \\
    WinoGrande & -- & Acc Char & 0 & -- \\
    Xsum & -- & ROUGE-2 & 1 & -- \\
    \hline
\end{longtable}

\subsection{Data: Human Evaluation Scores}
\label{app:sec:experimental_methodology:human_evals}

Human data annotators were hired to evaluate outputs of chat language models (LMs) in single-turn and multi-turn conversations using a Likert scale \citep{likert1932technique} from 1 to 7. The conversations were constructed within our taxonomy of areas-categories-subcategories (Sec. \ref{sec:methods}; Fig. \ref{fig:human_eval_prompt_taxonomy}). Each conversation was evaluated by at least three unique humans for a combined total of 2104 unique human annotators. Our human annotators scored 11291 single-turn conversations and 2081 multi-turn conversations.

\begin{longtable}{|l|l|l|}
\caption{\textbf{Human Evaluation Areas, Categories, and Subcategories.}}
\label{tab:categories}\\
\hline
\textbf{Area} & \textbf{Category} & \textbf{Subcategory}\\
\hline
\endfirsthead
\multicolumn{3}{c}%
{{\bfseries \tablename\ \thetable{} -- continued from previous page}}\\
\hline
\textbf{Area} & \textbf{Category} & \textbf{Subcategory}\\
\hline
\endhead
Dialogue & Adversarial Dishonesty & Adversarial Dishonesty \\
Dialogue & Adversarial Harmfulness & Adversarial Harmfulness \\
Dialogue & Advice & Casual advice \& recommendations \\
Dialogue & Advice & Personal \& interpersonal relationships \\
Dialogue & Brainstorming & Brainstorming \\
Dialogue & Classification & Classification \\
Dialogue & Closed QA & Closed QA \\
Dialogue & Code & Code \\
Dialogue & Conversational / Entertainment & Conversational / Entertainment \\
Dialogue & Conversational/Entertainment & Conversational/Entertainment \\
Dialogue & Dialogue & Dialogue \\
Dialogue & Extraction & Extraction \\
Dialogue & Factual Questions & Factual Questions \\
Dialogue & Identity / Personas & Famous historical personalities \\
Dialogue & Identity / Personas & Fictional characters \\
Dialogue & Identity / Personas & No character (it's just an AI) \\
Dialogue & Identity / Personas & Public figures \\
Dialogue & Identity / Personas & Synthetic (made up) characters \\
Dialogue & Language Assistance & Language Assistance \\
Dialogue & Math & Math \\
Dialogue & Mathematical Reasoning & Mathematical Reasoning \\
Dialogue & Open QA & Open QA \\
Dialogue & Procedural Questions & Procedural Questions \\
Dialogue & Reasoning (math / problem-solving) & Reasoning (math / problem-solving) \\
Dialogue & Recommendations / Brainstorming & Recommendations / Brainstorming \\
Dialogue & Rewriting & Rewriting \\
Dialogue & Safety & Safety \\
Dialogue & Summarization & Summarization \\
Dialogue & Writing & Writing \\
Dialogue & Writing \& Content Creation & Writing \& Content Creation \\
Factual Questions & Cultural \& social topics & Arts \& literature \\
Factual Questions & Cultural \& social topics & History \& traditions \\
Factual Questions & Cultural \& social topics & Popular culture \& media \\
Factual Questions & Cultural \& social topics & Religion \& spirituality \\
Factual Questions & Cultural \& social topics & Social issues \& current events \\
Language assistance & Grammar, spelling, \& vocabulary & English slang \\
Language assistance & Grammar, spelling, \& vocabulary & Grammar \& syntax \\
Language assistance & Grammar, spelling, \& vocabulary & Language conventions \& style \\
Language assistance & Grammar, spelling, \& vocabulary & Spelling \& orthography \\
Language assistance & Grammar, spelling, \& vocabulary & Vocabulary \& word choice \\
Recommendations & Entertainment suggestions & Books, authors, \& genres \\
Recommendations & Entertainment suggestions & Games, apps, \& digital content \\
Recommendations & Entertainment suggestions & Movies, TV shows, \& streaming content \\
Recommendations & Entertainment suggestions & Music, songs, \& artists \\
Recommendations & Personal \& professional development & Health, wellness, \& self-improvement tips \\
Recommendations & Personal \& professional development & Job search \& career advice \\
Recommendations & Personal \& professional development & Networking \& mentorship opportunities \\
Recommendations & Personal \& professional development & Skill-building resources \& courses \\
Writing \& content creation & Creative writing & Articles \& reviews \\
Writing \& content creation & Creative writing & Fictional stories \& narrative \\
Writing \& content creation & Creative writing & In-the-style-of writing \\
Writing \& content creation & Creative writing & Poetry \& songwriting \\
Writing \& content creation & Creative writing & Social media posts \\
Writing \& content creation & Summarization \& editing & Content restructuring \& organization \\
Writing \& content creation & Summarization \& editing & Proofreading \\
Writing \& content creation & Summarization \& editing & Style \& tone adjustments \\
\hline
\end{longtable}


\subsection{Analyses: Linear Regression}
\label{app:sec:generalization_of_overparameterized_models}

Due to space limitations in the main text, we defered citations regarding generalization of overparameterized models to here. For a nonexhaustive list, please see \citet{vallet1989hebb,krogh1991simple,geman1992neural,krogh1992generalization,opper1995statistical,duin2000classifiers, spigler2018jamming, belkin2019reconciling, bartlett2020benign, belkin2020twomodels, nakkiran2021deep, poggio2019double, advani2020high, liang2020just, adlam2020understanding, rocks2022memorizing, rocks2021geometry, rocks2022bias, mei2022generalization, hastie2022surprises, bach2023highdimensional,schaeffer2023double, schaeffer2023divergence, curth2024u, schaeffer2024doubledescentdemystified}.

\clearpage
\section{Statistics of Human Evaluations}

\begin{figure}[h!]
    \centering
    \includegraphics[width=0.48\linewidth]{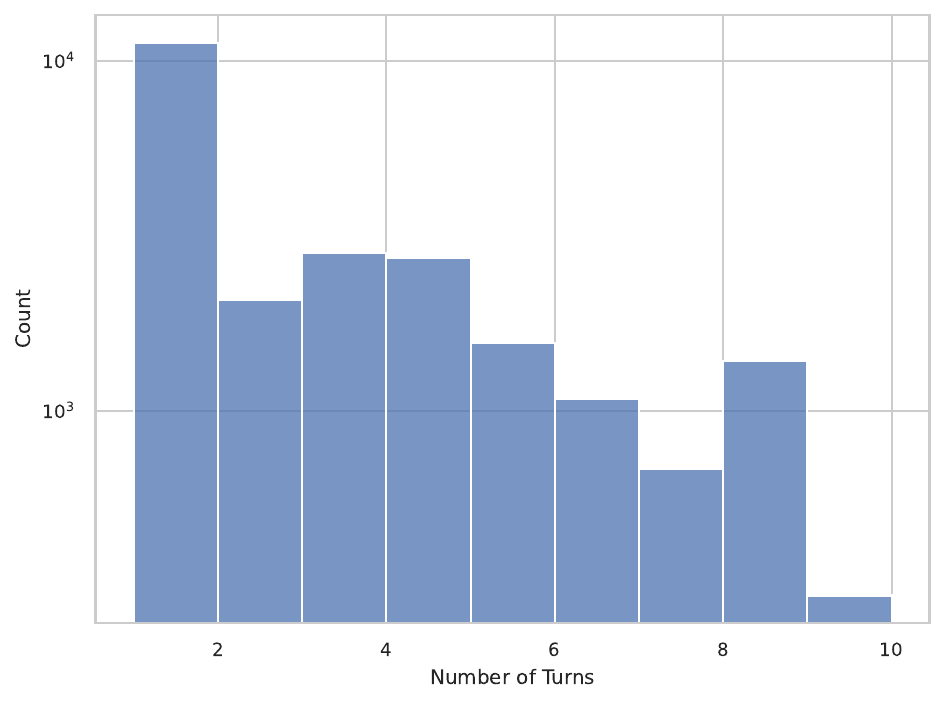}%
    \includegraphics[width=0.48\linewidth]{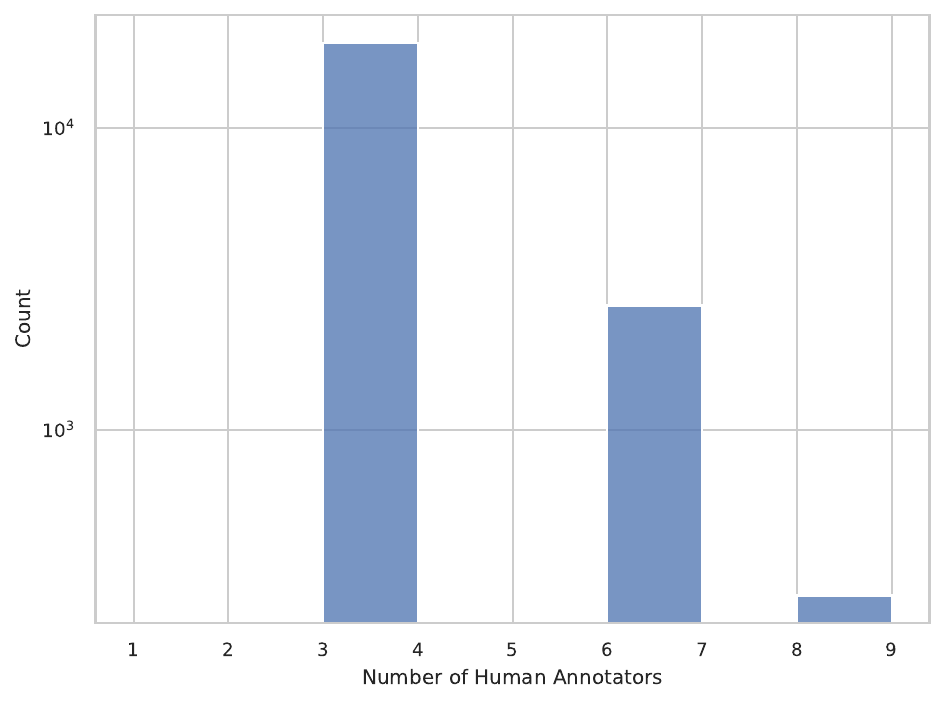}
    \caption{\textbf{Statistics of Human-Evaluated Data.} Left: Number of turns per datum. Right: Number of human annotators per human-evaluated datum.}
    \label{app:fig:human_evaluations_number_of_turns_per_annotated_sample}
\end{figure}

\begin{figure}[h!]
    \centering
    \includegraphics[width=0.48\linewidth]{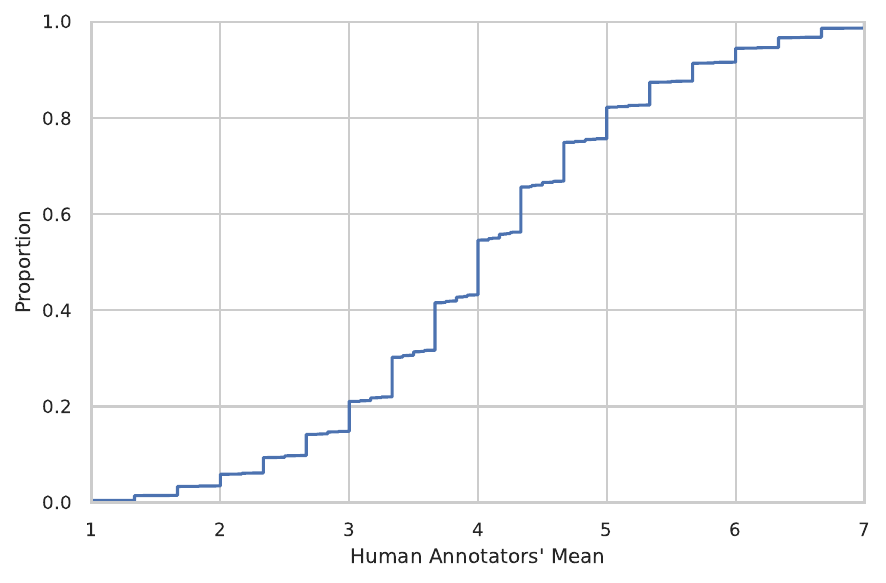}%
    \includegraphics[width=0.48\linewidth]{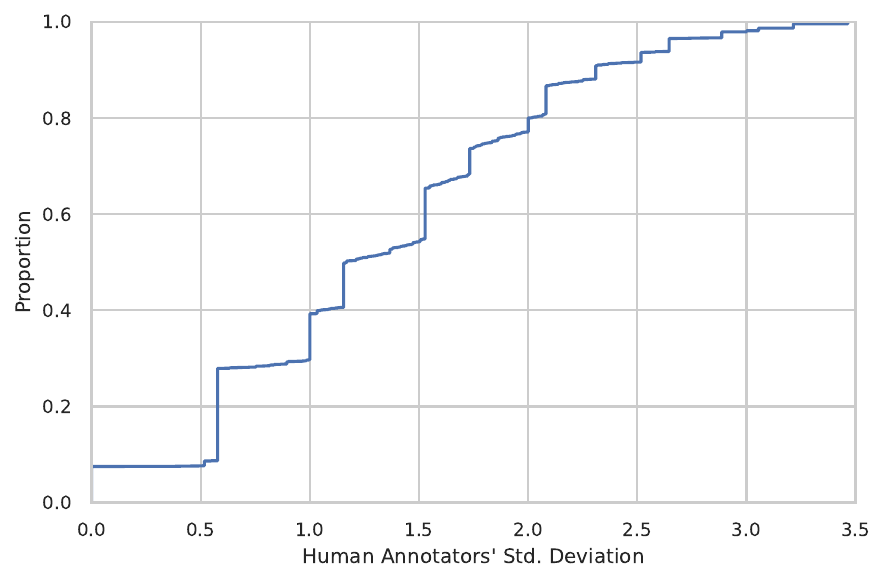}
    \caption{\textbf{Empirical Cumulative Distribution Functions of Human Annotators' Scores.} Left: Average of human annotators' score per annotated sample. Right: Human annotators' standard deviation per annotated sample. }
    \label{app:fig:human_annotator_scores_mean_std_dev_ecdfs}
\end{figure}

\begin{figure}[h!]
    \centering
    \includegraphics[width=0.53\linewidth]{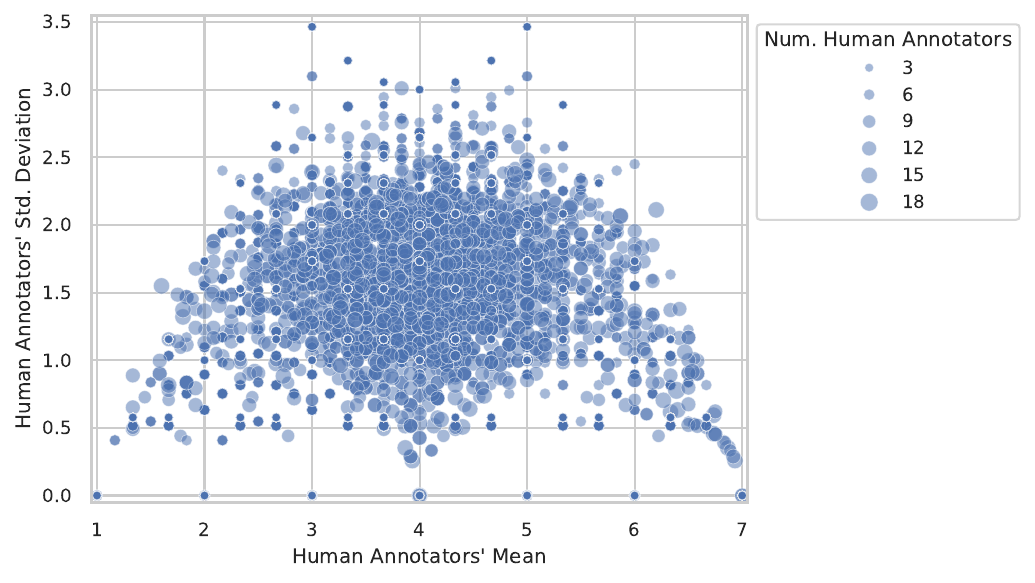}%
    \includegraphics[width=0.45\linewidth]{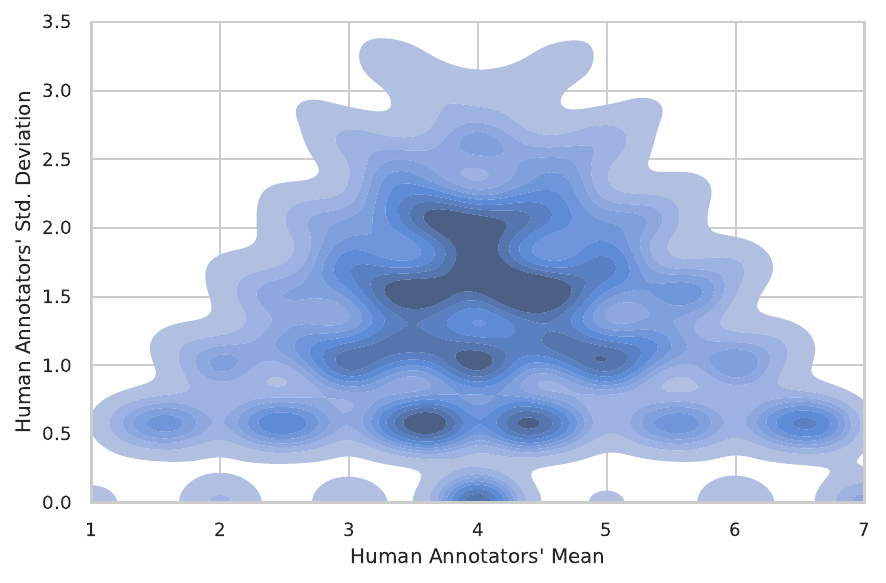}
    \caption{\textbf{Joint Distribution of Human Annotators' Average Scores per Datum vs Standard Deviation per Datum.} Left: Scatterplot. Right: Kernel Density Estimate.}
    \label{app:fig:human_annotator_scores_means_vs_std_dev_scatter_and_kde}
\end{figure}

As exploratory data analysis, we calculated and examined basic statistics of the human evaluations. Fig. \ref{app:fig:human_evaluations_number_of_turns_per_annotated_sample} showcases how many turns (i.e., back and forth messages) are in each sample evaluated by human annotators (left)  and how many human annotators evaluated each sample (right). Fig. \ref{app:fig:human_annotator_scores_mean_std_dev_ecdfs} shows the empirical cumulative distributions functions of the average of human annotators' scores per datum (left) and the standard deviation of human annotators' scores per datum (right). Fig. \ref{app:fig:human_annotator_scores_means_vs_std_dev_scatter_and_kde} visualizes the joint distribution of means and standard deviations of human annotators' scores per datum as both a scatterplot (left) and a kernel density estimate (right).

\clearpage
\section{Correlation Metrics Are Themselves Highly Correlated}
\label{app:sec:comparison_of_correlation_methods}

\begin{figure}[h!]
    \centering
    \includegraphics[width=\textwidth]{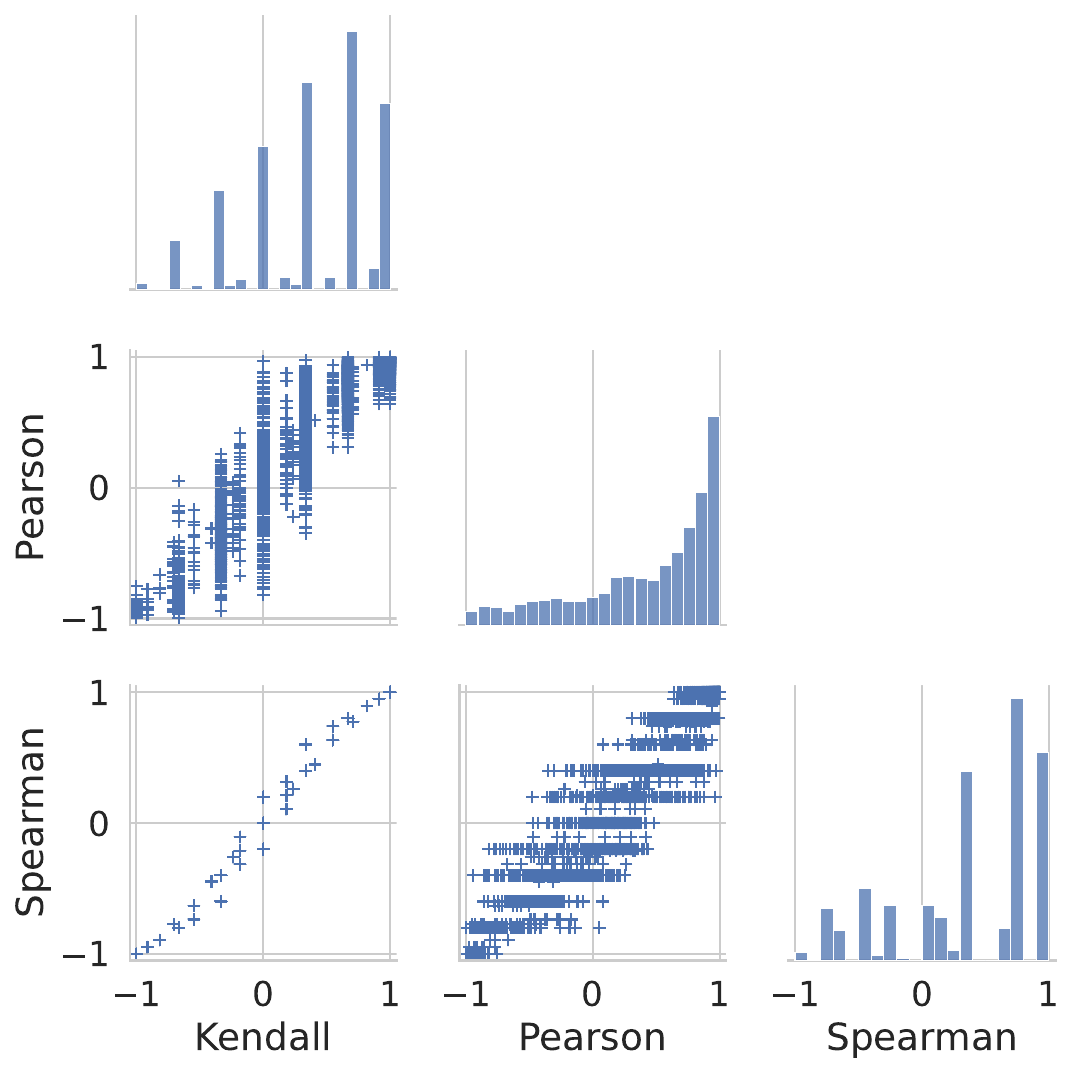}
    \caption{\textbf{Correlation of Academic-Human Evaluation Correlations Under Different Correlation Metrics}. For each pair of human evaluation (area and category) and NLP benchmark (benchmark and subset), we computed the correlation between scores under one of 3 correlation metrics: Pearson, Spearman and Kendall. We then looked at how correlated the correlation scores under the 3 correlation metrics are. In general, all 3 are correlation metrics yield correlated scores. This demonstrates that the choice of correlation metric is relatively less important.}
    \label{app:fig:correlation_couplings}
\end{figure}

\clearpage
\section{Correlation Matrices Between Human Evaluations and NLP Benchmarks and Their Singular Value Decompositions}
\label{app:sec:full_correlation_matrices}

\begin{figure}[h!]
    \centering
    \includegraphics[width=0.9\textwidth]{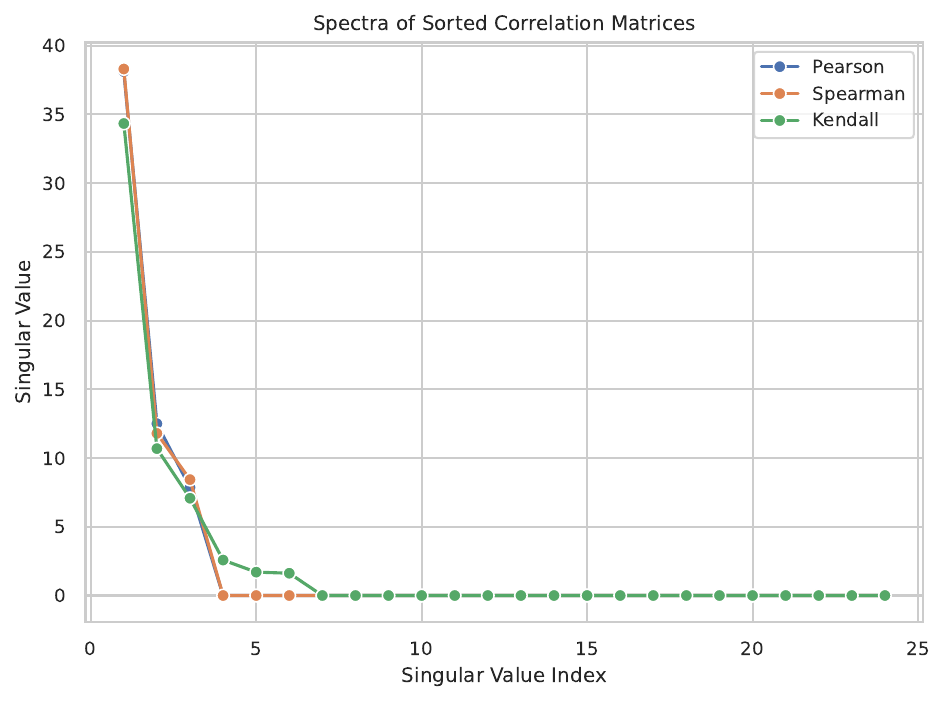}
    \caption{\textbf{Spectra of Human Evaluation-NLP Benchmark Correlation Matrices}. Because the correlation matrices are computed over four Chat Llama 2 models, the maximum matrix rank is 4. Howeer, both Pearson and Spearman correlation matrices have only 3 non-zero singular values.}
    \label{app:fig:academic_human_singular_value_spectra}
\end{figure}

\clearpage

\begin{figure}
    \centering
    \includegraphics[width=\textwidth]{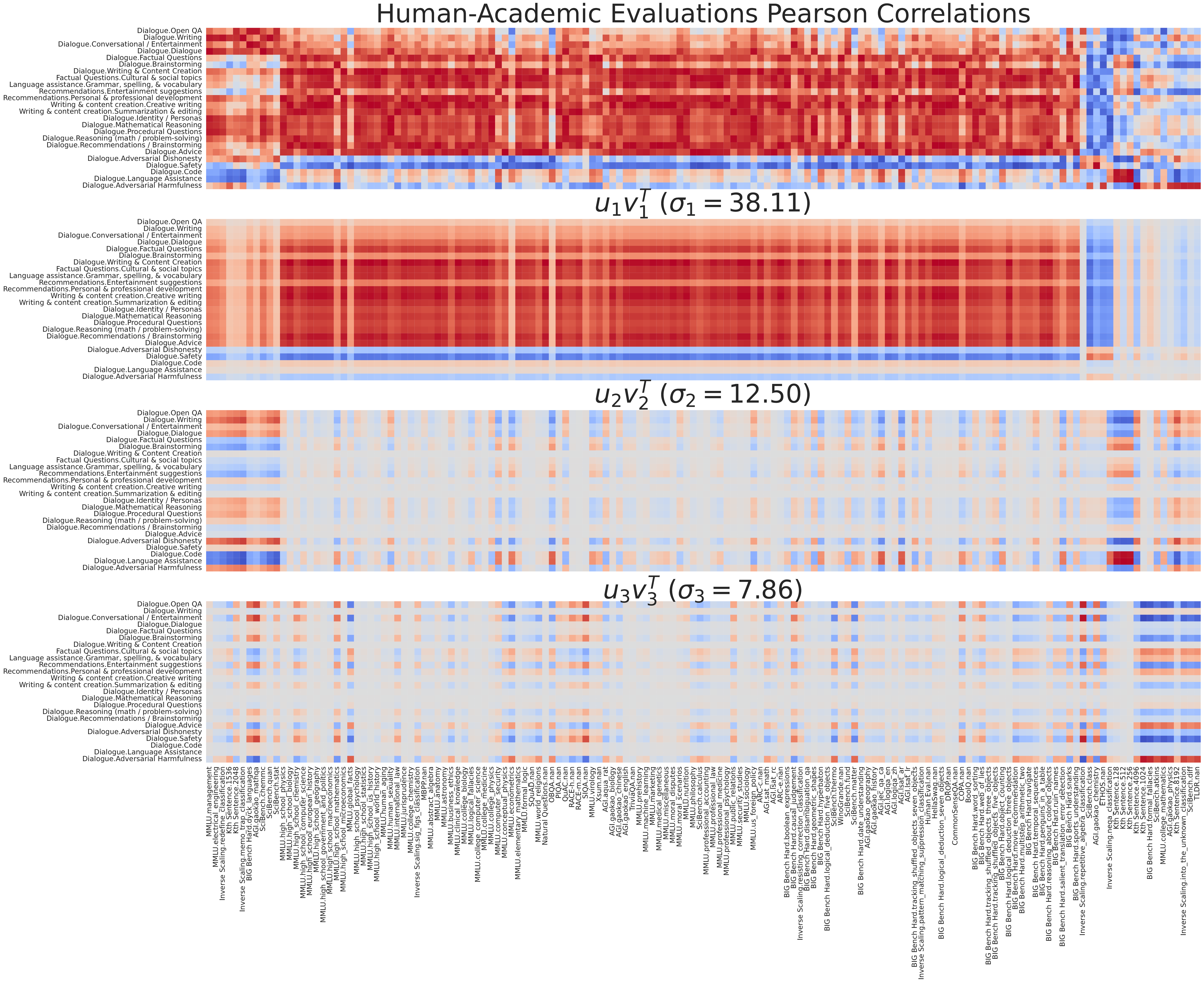}
    \caption{\textbf{Pearson Correlation Matrix between Human Evaluations and NLP Benchmarks.} The Pearson correlation matrix has 3 non-zero singular values, with corresponding modes shown in the last 3 rows.}
    \label{app:fig:pearson_correlation}
\end{figure}

\begin{figure}
    \centering
    \includegraphics[width=\textwidth]{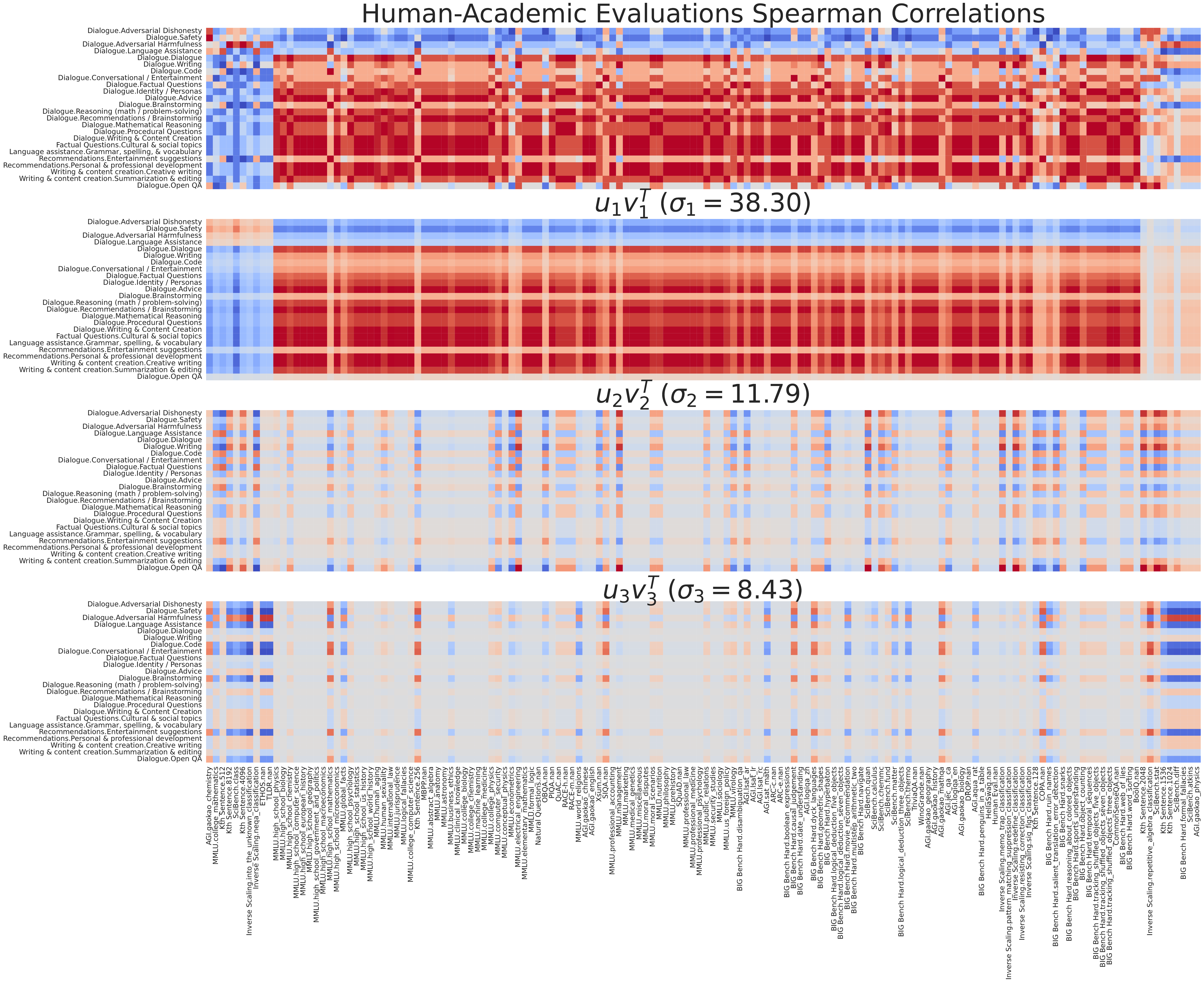}
    \caption{\textbf{Spearman Correlation Matrix between Human Evaluations and NLP Benchmarks.} The Spearman correlation matrix also has 3 non-zero singular values, with corresponding modes shown in the last 3 rows.}
    \label{app:fig:spearman_correlation}
\end{figure}

The first component of the Pearson correlation matrix divides the human evaluations and NLP benchmarks into 3 groups (Fig. \ref{fig:corr:singular_modes}B): one group that is broadly uncorrelated, and two unequally-sized groups that are self-correlated and mutually anti-correlated. The \textit{uncorrelated group} consists of human evaluations Dialogue:Code and Dialogue:Language Assistance, as well as NLP benchmarks Kth-sentence, TLDR, SciBench's Atkins and Differential Equations, MMLU's College Math and BBH's Formal Fallacies. The \textit{smaller self-correlated group} consists of Dialogue:Adversarial Dishonesty and Safety:Harmlessness as well as ETHOS (a hate speech detection benchmark), Inverse Scaling NEQA (a negation question-answering benchmark) and AGI Gaokao Chemistry, whereas the \textit{larger self-correlated group} consists of almost all other human evaluations and NLP benchmarks. This is more clearly visually displayed in the Spearman correlation matrix (App. Fig. \ref{app:fig:spearman_correlation}).


\clearpage

\clearpage

\begin{figure}
    \centering
    \includegraphics[width=\textwidth]{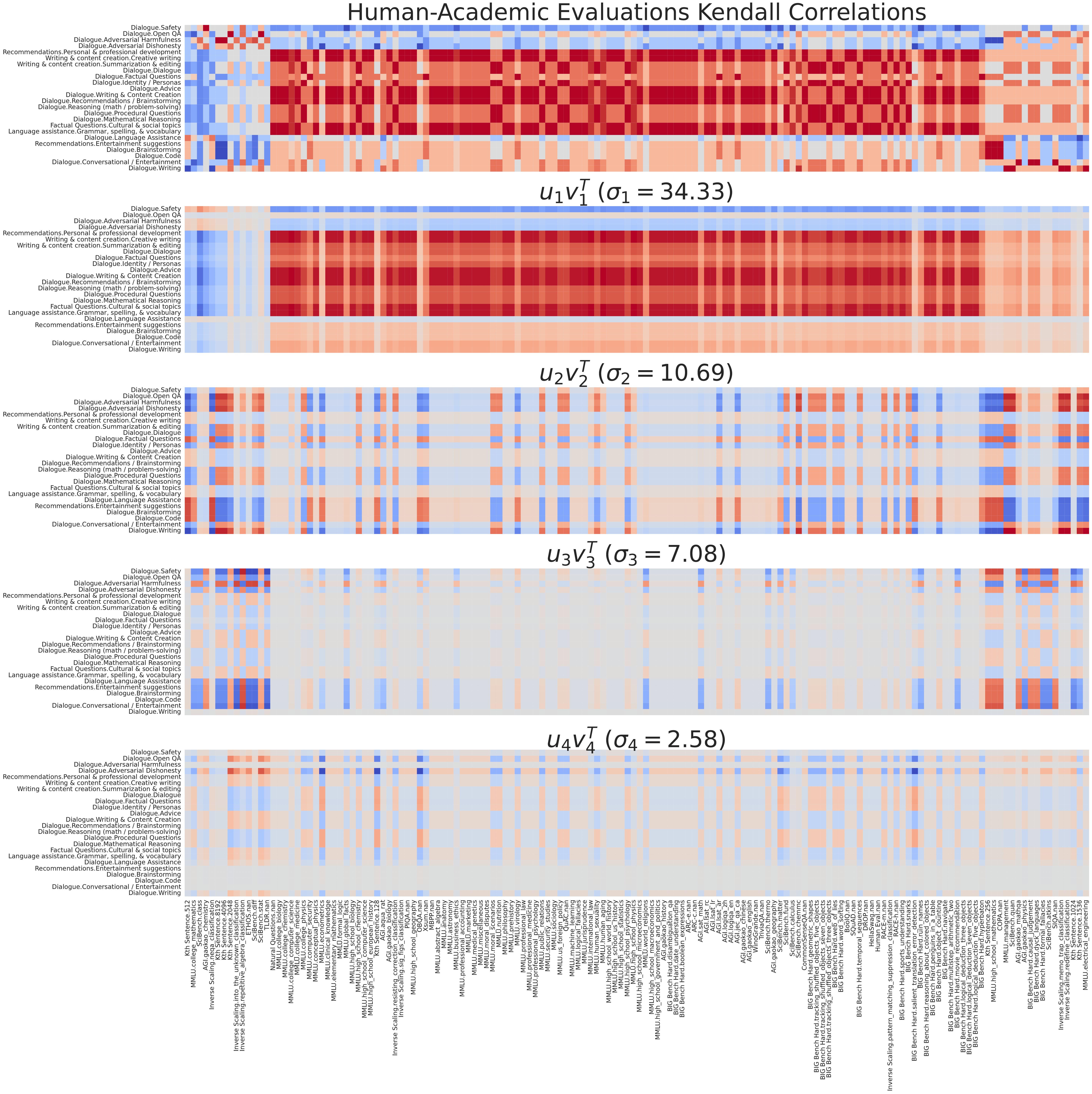}
    \caption{\textbf{Kendall Correlation Matrix between Human Evaluations and NLP Benchmarks.} The Kendall correlation matrix has four non-zero singular values, with corresponding modes shown in the last four rows.}
    \label{app:fig:kendall_correlation}
\end{figure}

\clearpage
\section{Empirical Scaling Behavior of Human Evaluations and NLP Benchmarks}
\label{app:empirical_scaling_behavior}

\begin{figure}[h]
    \centering
    \includegraphics[width=\textwidth]{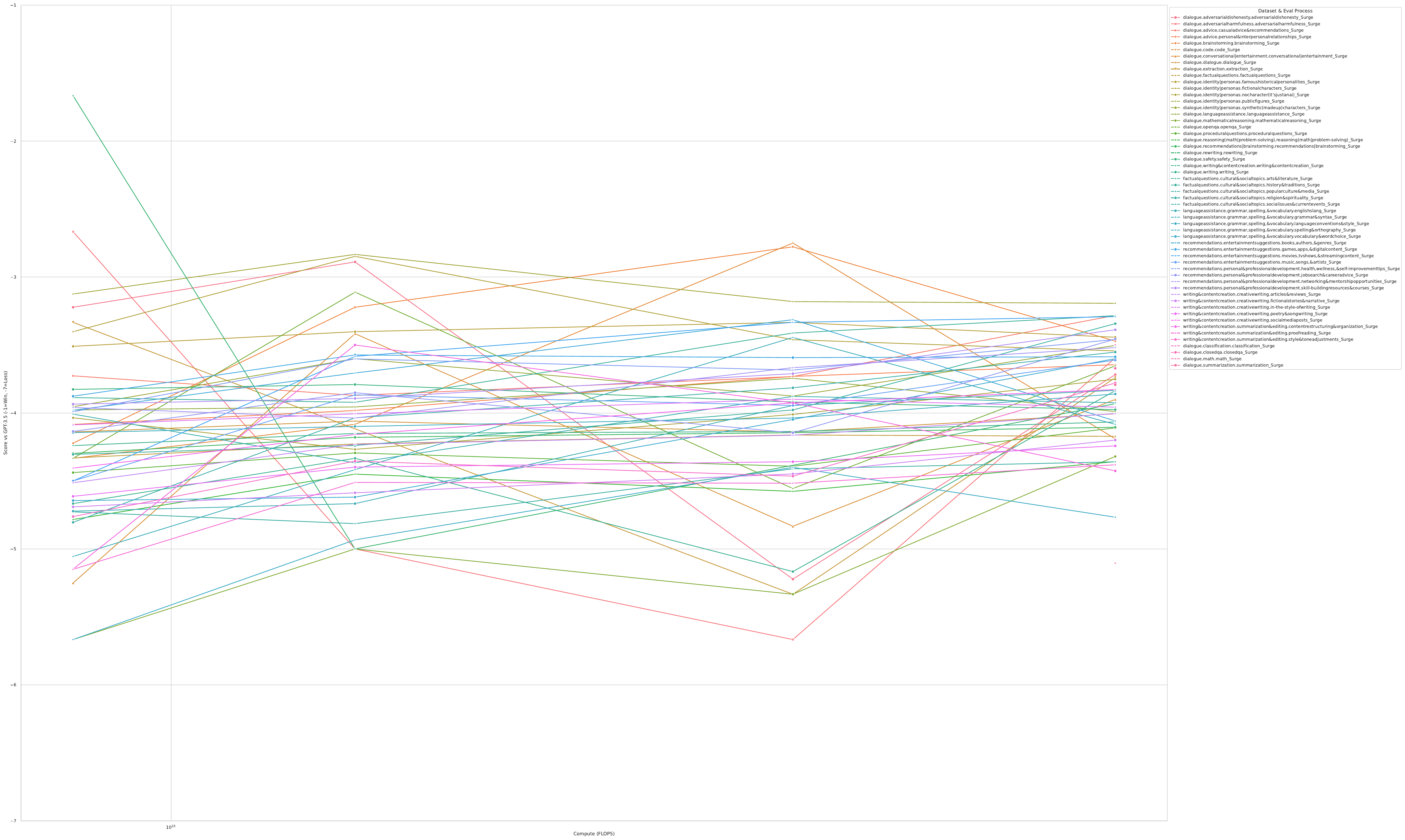}
    \caption{\textbf{Empirical Scaling Behavior of Human Evaluations with Increasing Compute.}}
    \label{app:fig:human_eval_vs_compute}
\end{figure}

\begin{figure}
    \centering
    \includegraphics[width=\textwidth]{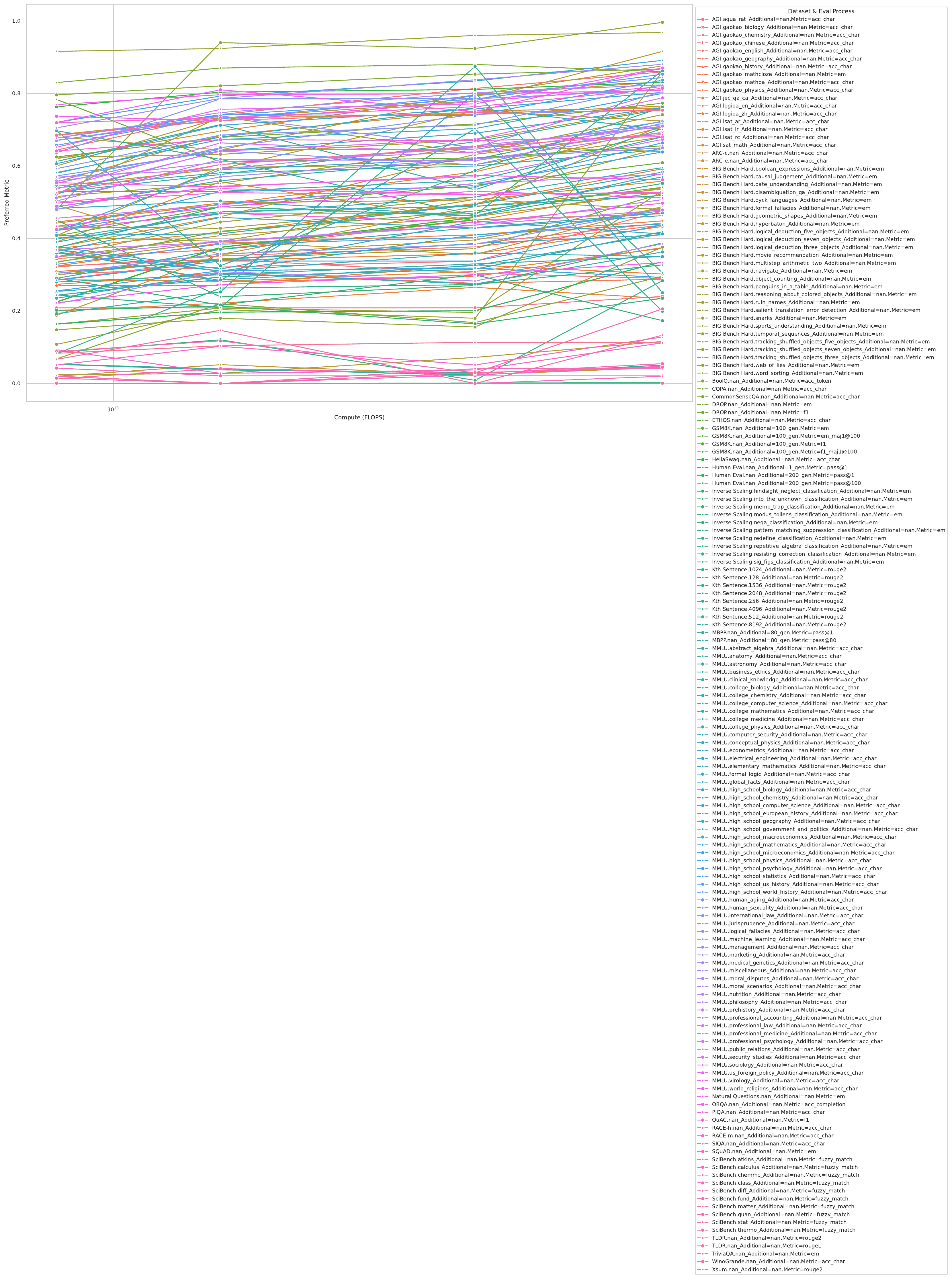}
    \caption{\textbf{Empirical Scaling Behavior of NLP Benchmarks with Increasing Compute.}}
    \label{app:fig:nlp_benchmark_vs_compute}
\end{figure}

\clearpage
\section{Coefficients of Leave-One-Out Cross-Validated Linear Regressions}

\begin{figure}[h!]
    \centering
    \includegraphics[width=\textwidth]{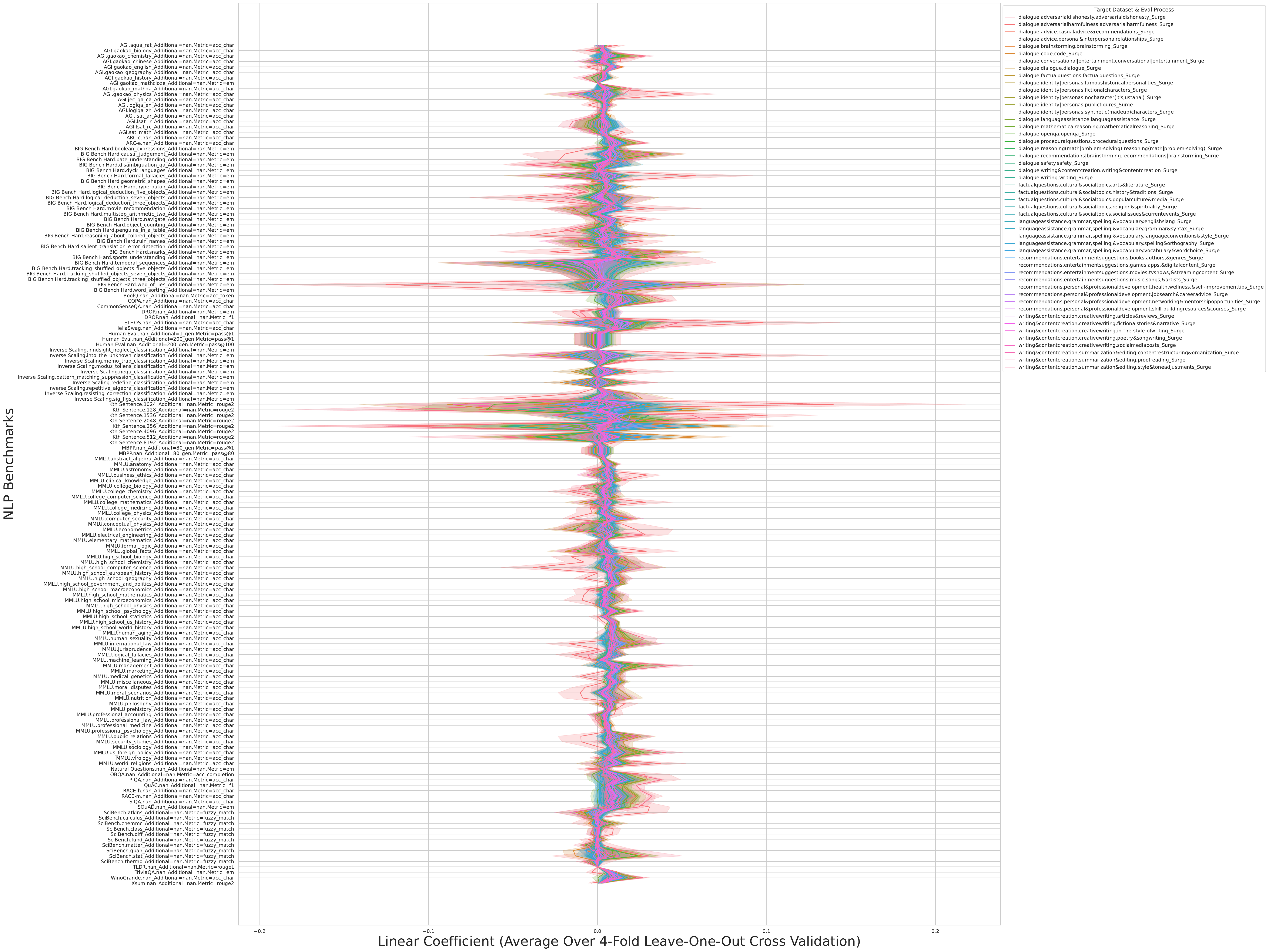}
    \caption{\textbf{Linear Coefficients of NLP Benchmarks in Predicting Human Evaluations.} For each human evaluation area, category and subcategory, we visualize the learned linear parameters per NLP benchmark averaged over the 4-fold leave-one-out cross validation process. For interpretation of results, see Sec. \ref{sec:predictions}.}
    \label{app:fig:linear_regression_coefficients}
\end{figure}

\end{document}